\documentclass[conference]{IEEEtran}

\usepackage{epsfig}
\usepackage{graphicx}
\usepackage{amsmath}
\usepackage{amssymb}
\usepackage{graphics} % for pdf, bitmapped graphics files
\usepackage{mathptmx} % assumes new font selection scheme installed
\usepackage{booktabs}
\usepackage{multirow}
\usepackage{color}
\usepackage{mathrsfs}
\usepackage{float}
\usepackage{subfigure}

\ifCLASSINFOpdf
\else
\fi

\begin{document}
\title{Monocular Depth Estimation Based On Deep Learning: An Overview}
%\title{Monocular Depth Estimation Based on Deep Learning: An Overview}

\author{\IEEEauthorblockN{Chaoqiang Zhao, \quad Qiyu Sun, \quad Chongzhen Zhang, \quad Yang Tang$^{*}$, \quad Feng Qian}
%\IEEEauthorblockA{School of Electrical and\\Computer Engineering\\
%Georgia Institute of Technology\\	
\IEEEauthorblockA{ East China University of Science and Technology, Shanghai, China, 200237\\
$^{*}$ Corresponding auther (email: yangtang@ecust.edu.cn)}
%Atlanta, Georgia 30332--0250\\
%Email: http://www.michaelshell.org/contact.html}
%\and
%\IEEEauthorblockN{Qiyu Sun}
%\IEEEauthorblockA{Twentieth Century Fox\\
%\and
%\IEEEauthorblockN{Chongzhen Zhang}
%\and
%\IEEEauthorblockN{Yang Tang$^{*}$}
%\and
%\IEEEauthorblockN{Feng Qian}
%\IEEEauthorblockA{Starfleet Academy\\
%San Francisco, California 96678--2391\\
}


\maketitle

\begin{abstract}
	Depth information is important for autonomous systems to perceive environments and estimate their own state.
Traditional depth estimation methods, like structure from motion and stereo vision matching, are built on feature correspondences of multiple viewpoints. Meanwhile, the predicted depth maps are sparse.
Inferring depth information from a single image (monocular depth estimation) is an ill-posed problem.
With the rapid development of deep neural networks, monocular depth estimation based on deep learning has been widely studied recently and achieved promising performance in accuracy.
Meanwhile, dense depth maps are estimated from single images by deep neural networks in an end-to-end manner.
In order to improve the accuracy of depth estimation, different kinds of network frameworks, loss functions and training strategies are proposed subsequently.
Therefore, we survey the current monocular depth estimation methods based on deep learning in this review.
%In this review, we focus on the related works in monocular depth estimation based on deep learning.
Initially, we conclude several widely used datasets and evaluation indicators in deep learning-based depth estimation.
Furthermore, we review some representative existing methods according to different training manners: supervised, unsupervised and semi-supervised.
%	Meanwhile, the novel frameworks and loss functions are briefly summarized in this review.
Finally, we discuss the challenges and provide some ideas for future researches in monocular depth estimation.
\end{abstract}

\IEEEpeerreviewmaketitle

%%%%%%%%% BODY TEXT
\section{Introduction}

Estimating depth information from images is one of the basic and important tasks in computer vision, which can be widely used in simultaneous localization and mapping (SLAM) \cite{hu2012robust}, navigation \cite{zhu2015vision}, object detection \cite{chai2017obstacle} and semantic segmentation \cite{park2017rdfnet}, etc.

\textbf{Geometry-based methods:} Recovering 3D structures from a couple of images based on geometric constraints is popular way to perceive depth and has been widely investigated in recent forty years.
\textit{Structure from motion (SfM)} \cite{ullman1979interpretation} is a representative method for estimating 3D structures from a series of 2D image sequences and is applied in 3D reconstruction \cite{mancini2013using} and SLAM \cite{mur2015orb} %\Authorfootnote% 首页Email地址
successfully. The depth of sparse features are worked out by SfM through feature correspondences and geometric constraints between image sequences, i.e., the accuracy of depth estimation relies heavily on the exact feature matching and high-quality image sequences. Furthermore, SfM suffers from monocular scale ambiguity \cite{szeliski1997shape}.
\textit{Stereo vision matching} also has the ability to recover 3D structures of a scene by observing the scene from two viewpoints \cite{zou2010method,cao2015summary}. Stereo vision matching simulates the way of human eyes by two cameras, and the disparity maps of images are calculated through a cost function. Since the transformation between two cameras is calibrated in advance, the scale information is included in depth estimation during the stereo vision matching process, which is different from the SfM process based on monocular sequences \cite{benosman1996multidirectional,ramirez2020improve}.

\begin{figure}[t]
	\centering
	\includegraphics[width = \columnwidth]{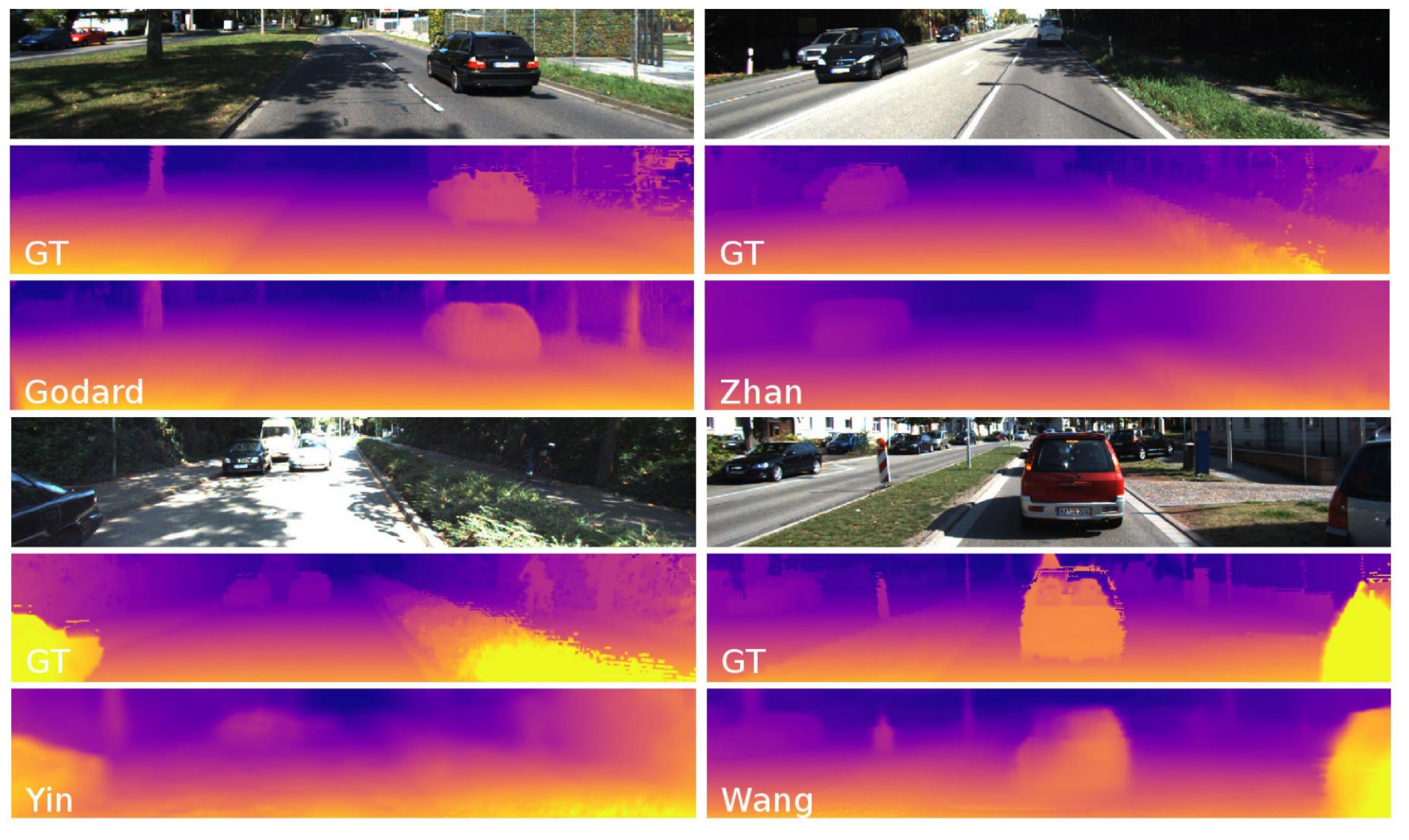}
	%\caption{fig1}
	\caption{An example of monocular depth estimation. The depth maps are predicted from deep neural networks proposed by Godard \textit{et al.} \cite{godard2017unsupervised}, Zhan \textit{et al.} \cite{zhan2018unsupervised}, Yin \textit{et al.} \cite{yin2018geonet}, and Wang \textit{et al.} \cite{wang2018learning}. The results are taken from \cite{fei2019geo}. As shown in these figures, the 3D structures of objects, like trees, street and cars, can be effectively perceived from single images by deep depth networks.}
	\label{fig:fig1}
\end{figure}

Although the above geometry-based methods can efficiently calculate the depth values of sparse points, these methods usually depend on image pairs or image sequences \cite{mancini2013using,cao2015summary}. How to get the dense depth map from a single image is still a significant challenge because of lack of effective geometric solutions.

\textbf{Sensor-based methods:} Utilizing depth sensors, like RGB-D cameras and LIDAR, is able to get the depth information of the corresponding image directly.
RGB-D cameras have the ability to get the pixel-level dense depth map of RGB image directly, but they suffer from the limited measurement range and outdoor sunlight sensitivity \cite{tateno2017cnn}.
Although LIDAR is widely used in unmanned driving industry for depth measurement \cite{yoneda2014lidar}, it can only generate the sparse 3D map.
Besides, the large size and power consumption of these depth sensors (RGB-D cameras and LIDAR) affect their applications to small robotics, like drones.
Due to the low cost, small size and wide applications of monocular cameras, estimating the dense depth map from a single image has received more attention, and it has been well researched based on deep learning in an end-to-end manner recently.

\textbf{Deep learning-based methods:} With the rapid development in deep learning, deep neural networks show their outstanding performance on image processing, like image classification \cite{zhang2019fast}, objective detection \cite{pang2019libra} and semantic segmentation \cite{lyu2019esnet}, etc, and related well-written overviews can be found in \cite{zhao2019object,ghosh2019understanding,rawat2017deep,tang2020perception}. Besides, recent developments have shown that the pixel-level depth map can be recovered from a single image in an end-to-end manner based on deep learning \cite{facil2019cam}. A variety of neural networks have manifested their effectiveness to address the monocular depth estimation, such as convolutional neural networks (CNNs) \cite{garg2016unsupervised}, recurrent neural networks (RNNs) \cite{wang2019recurrent}, variational auto-encoders (VAEs) \cite{chakravarty2019gen} and generative adversarial networks (GANs) \cite{aleotti2018generative}. The main goal of this overview is to provide an intuitive understanding of mainstream algorithms that have made significant contributions to monocular depth estimation. We review some related works in monocular depth estimation from the aspect of learning methods, including the loss function and network framework design, which is different from our previous review \cite{tang2020perception}. Some examples of monocular depth estimation based on deep learning are shown in Fig. \ref{fig:fig1}.

This survey is organized in the following way: Section II introduces some widely used datasets and evaluation indicators in monocular depth estimation.
Section III reviews some representative depth estimation methods based on deep learning according to different training modes. We also conclude some novel frameworks that can effectively improve network performance.
Section IV summarizes the current challenges and promising directions to research.
Section V concludes this review.

%-------------------------------------------------------------------------

\section{Datasets and evaluation indicators in depth estimation}

\subsection{Datasets}

\textbf{KITTI:} The KITTI dataset \cite{geiger2012we} is the largest and most commonly used dataset for the sub-tasks in computer vision, like optical flow \cite{mayer2016large}, visual odometry \cite{zhao2019deep}, depth \cite{eigen2014depth}, object detection \cite{chen2017multi}, semantic segmentation \cite{wang2018understanding} and tracking \cite{chang2019argoverse}, etc. It is also the commonest benchmark and the primary training dataset in the unsupervised and semi-supervised monocular depth estimation.
The real images from ``city", ``residential" and ``road" categories are collected in the KITTI dataset, and the 56 scenes in the KITTI dataset are divided into two parts, 28 ones for training and the other 28 ones for testing, by Eigen \textit{et al.} \cite{eigen2014depth}. Each scene consists of stereo image pairs with a resolution of 1224$\times$368.
The corresponding depth of every RGB image is sampled in a sparse way by a rotating LIDAR sensor.
%Besides, the ground truth of depth is only available at the bottom of the image.
Since the dataset also provides the ground truth of pose for 11 odometry sequences, it is also widely used to evaluate deep learning-based visual odometry (VO) algorithms \cite{xue2019beyond,clark2017vinet}.

\textbf{NYU Depth:} The NYU Depth dataset \cite{silberman2012indoor} focuses on indoor environments, and there are 464 indoor scenes in this dataset. Different from the KITTI dataset, which collects ground truth with LIDAR, the NYU Depth dataset takes monocular video sequences of scenes and the ground truth of depth by an RGB-D camera.
It is the common benchmark and the primary training dataset in the supervised monocular depth estimation.
These indoor scenes are split into 249 ones for training and 215 ones for testing. The resolution of the RGB images in sequences is 640$\times$480, and they are also down-sampled by half during experiments. Because of the different variable frame rates between RGB camera and depth camera, it is not a one-to-one correspondence between depth maps and RGB images. In order to align the depth maps and the RGB images, each depth map is associated with the closest RGB image at first. Then, with the geometrical relationship provided by the dataset, the camera projections are used to align depth and RGB pairs. Since the projection is discrete, not all pixels have a corresponding depth value, and thus the pixels with no depth value are masked off during the experiments.

\textbf{Cityscapes:} The Cityscapes dataset \cite{cordts2016cityscapes} mainly focuses on semantic segmentation tasks \cite{wang2018understanding}. There are 5,000 images with fine annotations and 20,000 images with coarse annotations in this dataset. Meanwhile, this dataset consists of a set of stereo video sequences, which are collected from 50 cities for several months. Since this dataset does not contain the ground truth of depth, it is only applied to the training process of several unsupervised depth estimation methods \cite{godard2017unsupervised,yin2018geonet}. The performance of depth networks is improved by pre-training the networks on the Cityscapes, and the experiments in \cite{zhou2017unsupervised,yin2018geonet,godard2017unsupervised,bian2019unsupervised} have proved the effectiveness of this joint training method. The training data consists of 22,973 stereo image pairs with a resolution of 1024$\times$2048 collected from different cities.

\textbf{Make3D:} The Make3D dataset \cite{saxena2008make3d} only consists of monocular RGB as well as depth images and does not have stereo images, which is different from the above datasets. Since there are no monocular sequences or stereo image pairs in this dataset, semi-supervised and unsupervised learning methods do not use it as the training set, while supervised methods usually adopt it for training. Instead, it is widely used as a testing set of unsupervised algorithms to evaluate the generalization ability of networks on different datasets \cite{godard2017unsupervised}.

\subsection{Evaluation metrics}

%With the input of single images, the networks output the corresponding depth map.
In order to evaluate and compare the performance of various depth estimation networks, a commonly accepted evaluation method is proposed in \cite{eigen2014depth} with five evaluation indicators: \textbf{ RMSE, RMSE log, Abs Rel, Sq Rel, Accuracies}. These indicators are formulated as:

\begin{itemize}
	\item $\textbf{RMSE} = \sqrt{\frac{1}{|N|}\sum_{i\in N}\parallel d_{i}-d_{i}^{*} \parallel^{2}}$,
	
	\item $\textbf{RMSE log} = \sqrt{\frac{1}{|N|}\sum_{i\in N}\parallel \log (d_{i})- \log (d_{i}^{*}) \parallel^{2}}$,
	
	\item $\textbf{Abs Rel} = \frac{1}{|N|}\sum_{i\in N}\frac{\mid d_{i}-d_{i}^{*}\mid}{d_{i}^{*}}$,\quad
	\item $\textbf{Sq Rel} = \frac{1}{|N|}\sum_{i\in N}\frac{\parallel d_{i}-d_{i}^{*}\parallel^{2}}{d_{i}^{*}}$,
	
	\item \textrm{\textbf{Accuracies:} $\%$ of $d_{i}$ s.t. } $\max(\frac{d_{i}}{d_{i}^{*}}, \frac{d_{i}^{*}}{d_{i}}) = \delta < thr$,
\end{itemize}
where $d_{i}$ is the predicted depth value of pixel $i$, and $d_{i}^{*}$ stands for the ground truth of depth. Besides, $N$ denotes the total number of pixels with real-depth values, and $thr$ denotes the threshold.

\section{Monocular depth estimation based on deep learning}

%Estimating the depth map from a single image is an ill-posed problem.
Since humans can use priori information of the world, it is capable for them to perceive the depth information from a single image.
Inspired by this, previous works achieve single-image depth estimation by combining some prior information, like the relationship between some geometric structures (sky, ground, buildings) \cite{hoiem2005automatic}. With the convincing performance in image processing, CNNs have also demonstrated a strong ability to accurately estimate dense depth maps from single images \cite{eigen2014depth}.
\cite{dijk2019neural} investigates which kind of cues the depth networks should exploit for monocular depth estimation based on the four published methods (MonoDepth \cite{godard2017unsupervised}, SfMLearner \cite{zhou2017unsupervised}, Semodepth \cite{kuznietsov2017semi} and LKVOLearner \cite{wang2018learning}).

Deep neural networks can be regarded as a black box, and the depth network will learn some structural information for depth inference with the help of supervised signals.
However, one of the biggest challenges of deep learning is the lack of enough datasets with ground truth, which is expensive to acquire. Therefore, in this section, we review the monocular depth estimation methods from the aspect of using ground truth: supervised methods \cite{kendall2017end}, unsupervised methods \cite{mahjourian2018unsupervised} and semi-supervised methods \cite{kuznietsov2017semi}.
Although the training processes of the unsupervised and semi-supervised methods rely on monocular videos or stereo image pairs, the trained depth networks predict depth maps from single images during the testing. We summarize the existing methods from the aspect of their training data, supervised signals and contributions in Table \ref{Tab01}. We also collect the quantitative results of the unsupervised and semi-supervised algorithms evaluated on the KITTI dataset in Table \ref{Tab02}.

\subsection{Supervised monocular depth estimation}

\textbf{A basic model for supervised methods:} The supervisory signal of supervised methods is based on the ground truth of depth maps, so that monocular depth estimation can be regarded as a regressive problem \cite{eigen2014depth}. The deep neural networks are designed to predict depth maps from single images. The differences between the predicted and real depth maps are utilized to supervise the training of networks:
\textbf{$\mathcal{L}_{2}$ Loss}:
\begin{equation}
\mathcal{L}_{2}(d,d^{*}) = \frac{1}{N} \sum_{i}^{N}||d-d^{*}||_{2}^{2}, \label{eq:6}
%\end{array}{1}
\end{equation}
Therefore, depth networks learn the depth information of scenes by approximating the ground truth.

\textbf{Methods based on different architectures and loss functions:} To the best of our knowledge, Eigen \textit{et al.} \cite{eigen2014depth} firstly solve the monocular depth estimation problem by CNNs. The proposed architecture is composed of two-component stacks (the global coarse-scale network and the local fine-scale network) is designed in \cite{eigen2014depth} to predict the depth map from a single image in an end-to-end way. During the training process, they use the ground truth of depth $d^{*}$ as the supervised signals, and the depth network predicts the $\log$ depth as $\log d$. The training loss function is set as:
\begin{equation}
\mathcal{L}(d,d^{*}) = \frac{1}{N} \sum_{i}^{N} y_{i}^{2} - \frac{\lambda}{N^{2}}(\sum_{i}^{N} y_{i})^{2} ,\label{eq:1}
\end{equation}
where $y_{i}^{2} =  \log (d) - \log (d^{*})$. $\lambda$ refers to the balance factor and is set to 0.5. The coarse-scale network is trained at first, and then the fine-scale network is trained to refine the results by fixing the parameters of the coarse-scale network. The experiments show that the fine-scale network is effective to refine the depth map estimated by the coarse-scale network.
In \cite{eigen2015predicting}, Eigen \textit{et al.} propose a general multi-scale framework capable of dealing with the tasks such as depth map estimation, surface normal estimation, and semantic label prediction from a single image. For depth estimation, based on Eq. (\ref{eq:1}), an additional loss function is proposed to promote the local structural consistency:
\begin{equation}
\mathcal{L}_{s} = \frac{1}{N} \sum_{i}^{n} [ (\nabla_{x}D_{i})^{2} + (\nabla_{y}D_{i})^{2} ], \label{eq:5}
%\end{array}{1}
\end{equation}
where $D_{i} = \log (d_{i}) - \log (d_{i}^{*})$, and $\nabla$ is the vector differential operator. This function calculates the gradients of the difference between the predicted depth and the ground truth in the horizontal and vertical directions.
Similarly, considering that optical flow is successfully solved by CNN through supervised learning, Mayer \textit{et al.} \cite{mayer2016large} extend the optical flow networks to disparity and scene flow estimation.  A fully CNN framework for monocular depth estimation is proposed in \cite{shelhamer2015scene}, and then the proposed framework jointly optimizes the intrinsic factorization to recover the input image.
Inspired by the outstanding performance of ResNet \cite{he2016deep}, Laina \textit{et al.} \cite{laina2016deeper} introduce residual learning to learn the mapping relation between depth maps and single images, therefore their network is deeper than previous works in depth estimation with higher accuracy.
Besides, the fully-connected layers in ResNet are replaced by up-sampling blocks to improve the resolution of the predicted depth map. During the training process, they use the reverse Huber (Berhu) \cite{zwald2012berhu} as the supervised signal of depth network, which is also used in \cite{zhang2018joint} and achieves a better result than $\mathcal{L}_{2}$ loss (Eq. (\ref{eq:6})):\\
\textbf{Berhu Loss}:
\begin{equation}
\mathcal{L}_{Berhu}(d,d^{*}) = \left\{ \begin{array}{ll}
|d-d^{*}| & \textrm{if $|d-d^{*}|\leq c$},\\
\frac{(d-d^{*})^{2}+c^{2}}{2c} & \textrm{if $|d-d^{*}| > c$},
\end{array} \right. \label{eq:7}
%\end{array}{1}
\end{equation}
where $c$ is a threshold and set to $\frac{1}{5}max_{i}(|d-d^{*}|)$. If $|x| <c$, the Berhu loss is equal to $\mathcal{L}_{1}$, and the berHu loss is equal to $\mathcal{L}_{2}$ when $|x|$ is outside this range. Because of the deeper fully convolutional network and the improved loss function, this work achieves a better result than the previous works \cite{shelhamer2015scene,mayer2016large} with fewer parameters and training data.

Mancini \textit{et al.} \cite{mancini2016fast} focus on the application of depth estimation in obstacle detection. Instead of predicting the depth from a single image, the proposed fully CNN framework in \cite{mancini2016fast} uses both monocular image and corresponding optical flow to estimate an accurate depth map.
Chen \textit{et al.} \cite{chen2016single} tackle the challenge of perceiving the single-image depth estimation in the wild by exploring a novel algorithm. Different from using the ground truth of depth as the supervised signal, their networks are trained by the relative depth annotations.
The variant of Inception Module \cite{szegedy2015going} is also utilized in their framework to make the network deeper. Since the monocular view contains few geometric details, Kendall \textit{et al.} \cite{kendall2017end} design a deep learning framework to learn the structure of scenes from stereo image pairs. Besides, a disparity map instead of the depth map is predicted by the designed network, and the ground truth disparity is used for supervision:
\begin{equation}
%\begin{array}{1}
\mathcal{L}(I,I^{*}) = \frac{1}{N} \sum_{i}^{N} ||Dis_{i}-Dis_{i}^{*}||_{1}, \label{eq:8}
\end{equation}
where $Dis_{i}$ stands for the predicted disparity of pixel $i$, and $Dis_{i}^{*}$ is the corresponding ground truth. Considering the slow convergence and local optimal solutions caused by minimizing mean squared error in log-space during training, Fu \textit{et al.} \cite{fu2018deep} regard the monocular depth estimation as an ordinal regression problem. As the uncertainty of the predicted depth values increase along with the ground truth depth values, it is better to allow larger estimation errors when predicting larger depth values, which cannot be well solved by the uniform discretization (UD) strategy. Therefore, a spacing-increasing discretization (SID) strategy is proposed in \cite{fu2018deep} to discretize depth and optimize the training process. To improve the transportability of depth network on different cameras, Facil \textit{et al.} \cite{facil2019cam} introduce the camera model into the depth estimation network, improving the generalization capabilities of networks. Although the above methods achieve outstanding accuracy, a large number of parameters in these works limit the applications of their network in practice, especially on embedded systems. Hence, Woft \textit{et al.} \cite{wofk2019fastdepth} address this problem by designing a lightweight auto-encoder network framework. Meanwhile, the network pruning is applied to reduce computational complexity and improves real-time performance.

\begin{figure*}[htbp]
	\centering
	\subfigure[The framework of raw GAN;]{
		\includegraphics[width =0.6 \columnwidth]{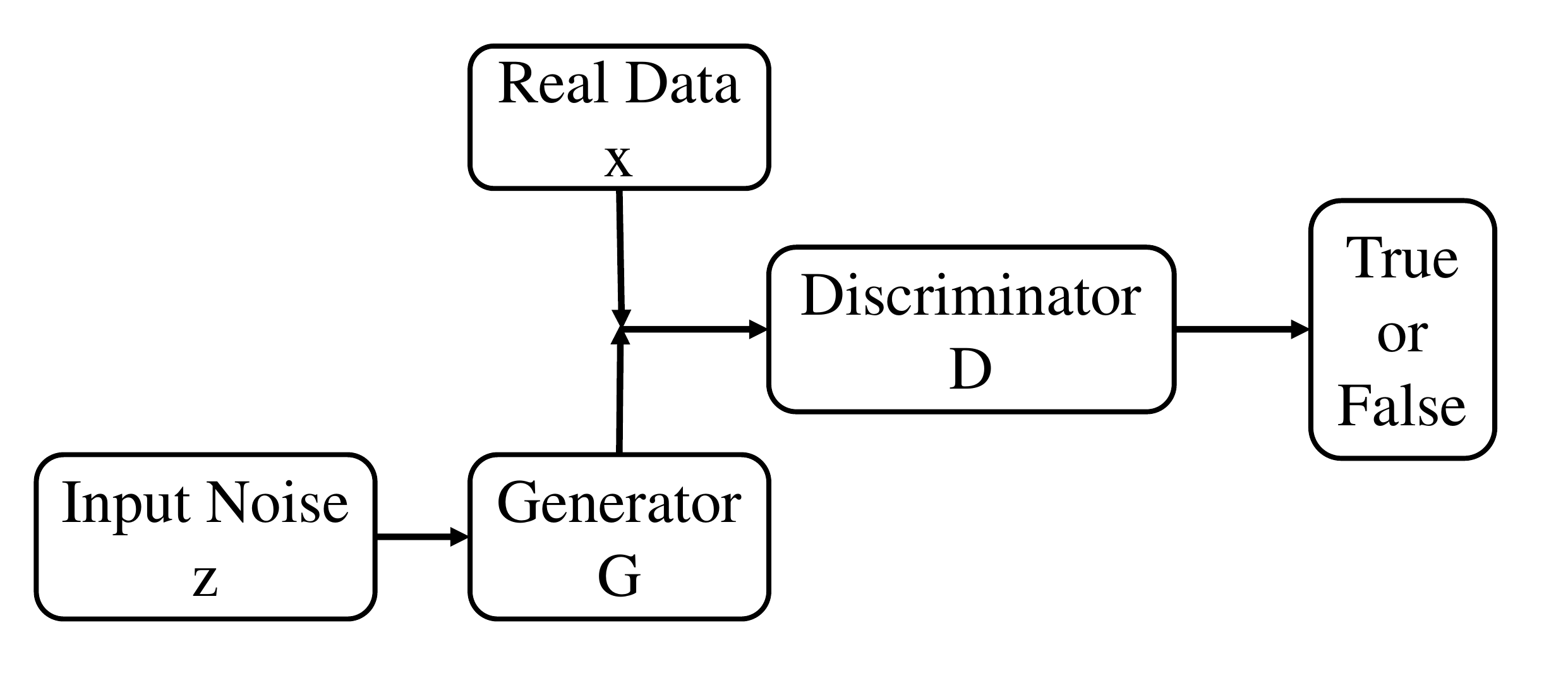}
		%\caption{fig1}
	}
	\subfigure[The framework of supervised methods based on GAN;]{
		\includegraphics[width =0.6 \columnwidth]{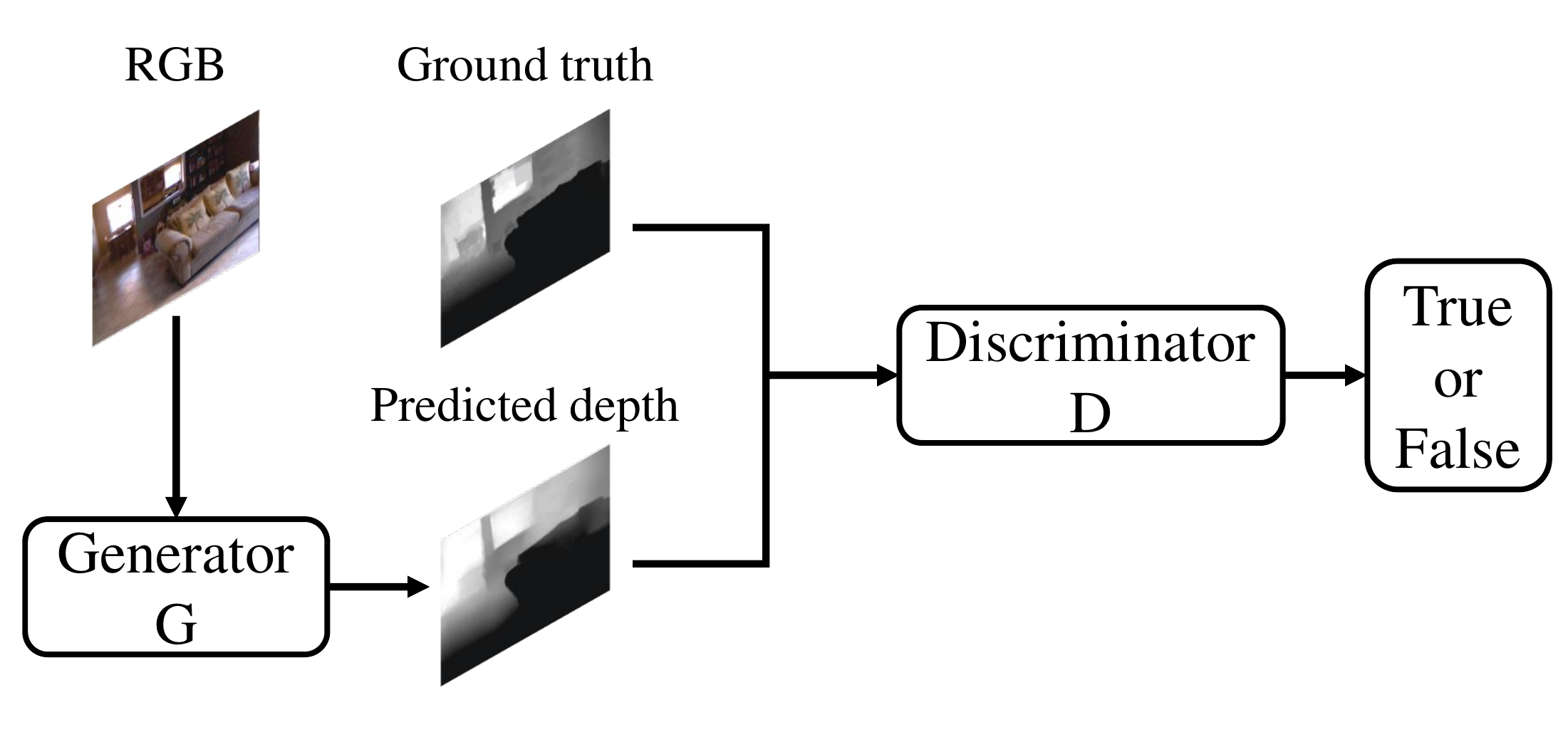}
	}
	\subfigure[The framework of unsupervised and semi-supervised methods based on GAN.]{
		\includegraphics[width = 0.6\columnwidth]{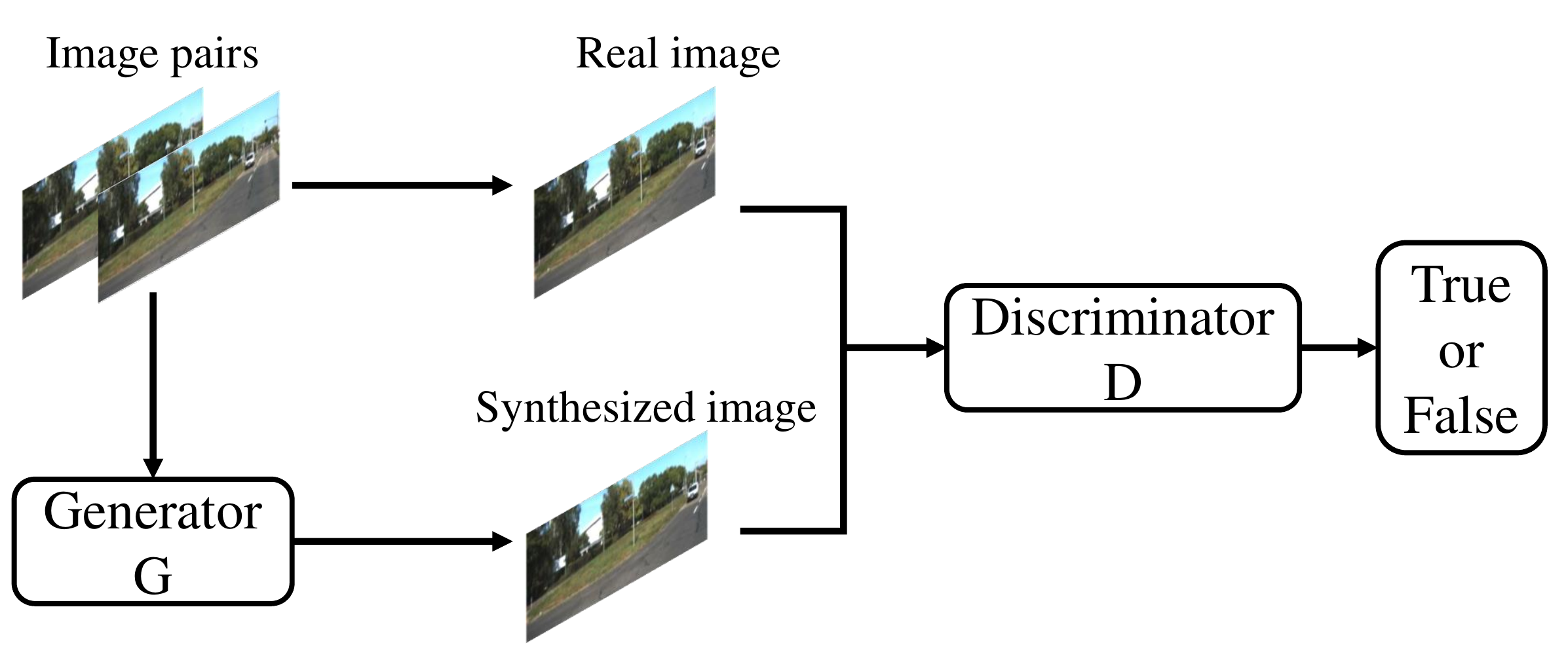}
	}
	\caption{ (a) The generator of raw GAN \cite{goodfellow2014generative} has the ability to generate the data from a vector $z$, and the discriminator is designed to distinguish the real and fake data. Finally, the data generated by generator has the same data distribution as the real data. (b) In the GAN-based supervised methods \cite{jung2017depth}, the depth maps predicted by generator (depth network) and real depth maps are sent to discriminator during training. (c) In the GAN-based unsupervised and semi-supervised methods \cite{cs2018monocular,feng2019sganvo}, owing to the lack of real dense depth maps, the RGB images synthesized by view reconstruction algorithm in generator and real images instead of depth maps are sent to discriminator, and the generator takes image pairs, like the image snippet in unsupervised methods or the stereo image pair in semi-supervised methods, to estimate the depth maps from single images and synthesize the RGB image.}
	\label{fig:fig3}
\end{figure*}

\textbf{Methods based on conditional random fields:} Instead of using an additional network to refine the results in \cite{eigen2014depth}, Li \textit{et al.} \cite{li2015depth} propose a refinement method based on the hierarchical conditional random fields (CRFs), which is also widely used for semantic segmentation \cite{huang2011hierarchical,ladicky2009associative}. Because of the continuous characteristics of depth between pixels, CRF can refine the depth estimation by considering the depth information of neighboring pixels, so that the CRF model is widely applied in the depth estimation \cite{li2015depth,liu2015learning,wang2015towards}. In \cite{li2015depth}, a deep CNN framework is designed to regress the depth map from multi-level image patches at the super-pixel level. Then, the depth map is refined from super-pixel level to pixel level via hierarchical CRF, and the energy function is:
\begin{equation}
\textbf{E(d)} = \sum_{i\in S} \phi_{i}(d_{i}) + \sum_{(i,j)\in \varepsilon_{s}} \phi_{ij} (d_{i},d_{j}) + \sum_{c\in P} \phi_{c}(d_{c}), \label{eq:2}
\end{equation}
where $S$ stands for the set of super-pixels, $\varepsilon_{s}$ refers to the set of super-pixel pairs that share a common boundary, and $P$ denotes the set of pixel-level patches. $\textbf{E(d)}$ consists of three parts: (i) a data term for calculating the quadratic distance between the depth value $d$ and the network regressed depth $\bar{d}$; (ii) a smoothness term for enforcing relevance between neighboring super-pixels; (iii) an auto-regression model for describing the local relevant structure in the depth map.
In the same year, a similar framework exploring deep CNN with continuous CRF, called deep convolutional neural fields, is proposed by Liu \textit{et al.} \cite{liu2015learning} to tackle the problem of monocular depth estimation. Meanwhile, a super-pixel pooling method is proposed by them to speed up the convolutional network, and it helps to design the deeper network to improve the accuracy of depth estimation.
Wang \textit{et al.} \cite{wang2015towards} present a framework jointly estimate the pixel-level depth map and semantic labels from a single image. Because of the structural consistency between the depth map and semantic labels, the interactions between the depth and semantic information are utilized to improve the performance of depth estimation. The depth and semantic prediction tasks are jointly trained by the supervised signal:
\begin{equation}
%\begin{array}{1}
\mathcal{L}(I,I^{*}) = \frac{1}{N} \sum_{i}^{N} (\log (d_{i}) - \log (d_{i}^{*}))^{2} - \lambda \frac{1}{N}\sum_{i}^{N} \log (P(l_{i}^{*})), \label{eq:3}
\end{equation}
\begin{equation}
P(l_{i}^{*}) = exp (z_{i,l_{i}^{*}})/ \sum_{l_{i}}exp (z_{i,l_{i}}), \label{eq:4}
%\end{array}{1}
\end{equation}
where $l_{i}^{*}$ stands for the ground truth of semantic labels. Meanwhile, $l_{i}$ refers to the predicted labels. $z_{i,l_{i}^{*}}$ denotes the output of the semantic node. To further refine the estimated depth, Wang \textit{et al.} also introduce a two-layer hierarchical CRF to update the depth details by extracting frequent templates for each semantic category, which lead to the fact that their methods cannot perform well as the number of classes increases.
Therefore, Mousavian \textit{et al.} \cite{mousavian2016joint} present a coupled framework for simultaneously estimating depth maps and semantic labels from a single image, and these two tasks share high-level feature representation of images extracted by CNN. A fully connected CRF is used and coupled with deep CNN to enhance the interactions between depth maps and semantic labels. Hence, their method is trained in an end-to-end manner and 10$x$ faster than that reported in \cite{wang2015towards}.
Zhang \textit{et al.} \cite{zhang2018joint} propose a task-attentional module to encapsulate the interaction and improve the performance of networks, which is different from previous works \cite{wang2015towards,mousavian2016joint}.
Similar to \cite{mousavian2016joint}, Xu \textit{et al.} \cite{xu2018structured} also integrate the continuous CRF model into deep CNN framework for end-to-end training. Besides, a structured attention model coupled with the CRF model is proposed in \cite{xu2018structured} to strengthen the information transfer between corresponding features.
The random forests (RFs) model is also introduced to monocular depth estimation tasks and efficiently enforces the accuracy of depth estimation \cite{roy2016monocular}.

\textbf{Methods based on adversarial learning:} Because of the outstanding performance on data generation \cite{zhang2017stackgan}, the adversarial learning proposed in \cite{goodfellow2014generative} has become a hot research direction in recent years. Varieties of algorithms, theories, and applications have been widely developed, which is reviewed in \cite{hong2019generative}. The frameworks of adversarial learning in depth estimation are shown in Fig. \ref{fig:fig3}. Different kinds of adversarial learning frameworks based on \cite{goodfellow2014generative}, like stacked GAN \cite{huang2017stacked}, conditional GAN \cite{mirza2014conditional} and Cycle GAN \cite{zhu2017unpaired}, are introduced into depth estimation tasks and have a positive impact on the depth estimation \cite{feng2019sganvo,jung2017depth,gwn2018generative}. In \cite{jung2017depth}, Jung \textit{et al.} introduce the adversarial learning into monocular depth estimation tasks. The generator consists of a Global Net and a Refinement Net, and these networks are designed to estimate the global and local 3D structures from a single image. Then, a discriminator is used to distinguish the predicted depth maps from the real ones, and this form is commonly used in supervised methods. The confrontation between generator $G$ and discriminator $D$ facilitates the training of the framework based on the min-max problem:
\begin{equation}
\mathop{\textrm{min}}\limits_{G} \mathop{\textrm{max}}\limits_{D} \mathbb{E}_{x \thicksim P_{gt}} [\log D(x)] + \mathbb{E}_{\hat{x} \thicksim P_{G}}[\log(1- D(\hat{x}))], \label{eq:9}
\end{equation}
where $x$ is the ground truth depth map, and $\hat{x}$ refers to the depth map predicted by generator.
Similarly, conditional GAN is also used in \cite{gwn2018generative} for monocular depth estimation.  The difference from \cite{jung2017depth} is that a secondary GAN is introduced to get a more refined depth map based on the image and coarse estimated depth map.

Because of being supervised by the ground truth, the supervised methods can effectively learn the functions to map 3D structures and their scale information from single images.
However, these supervised methods are limited by the labeled training sets, which are hard and expensive to acquire.

\subsection{Unsupervised monocular depth estimation}

Instead of using the ground truth,  which is expensive to acquire, the geometric constraints between frames are regarded as the supervisory signal during the training process of the unsupervised methods.

\begin{figure*}[htbp]
	\centering
	\subfigure[An illustration of image warping process;]{
		\includegraphics[width = 0.8\columnwidth]{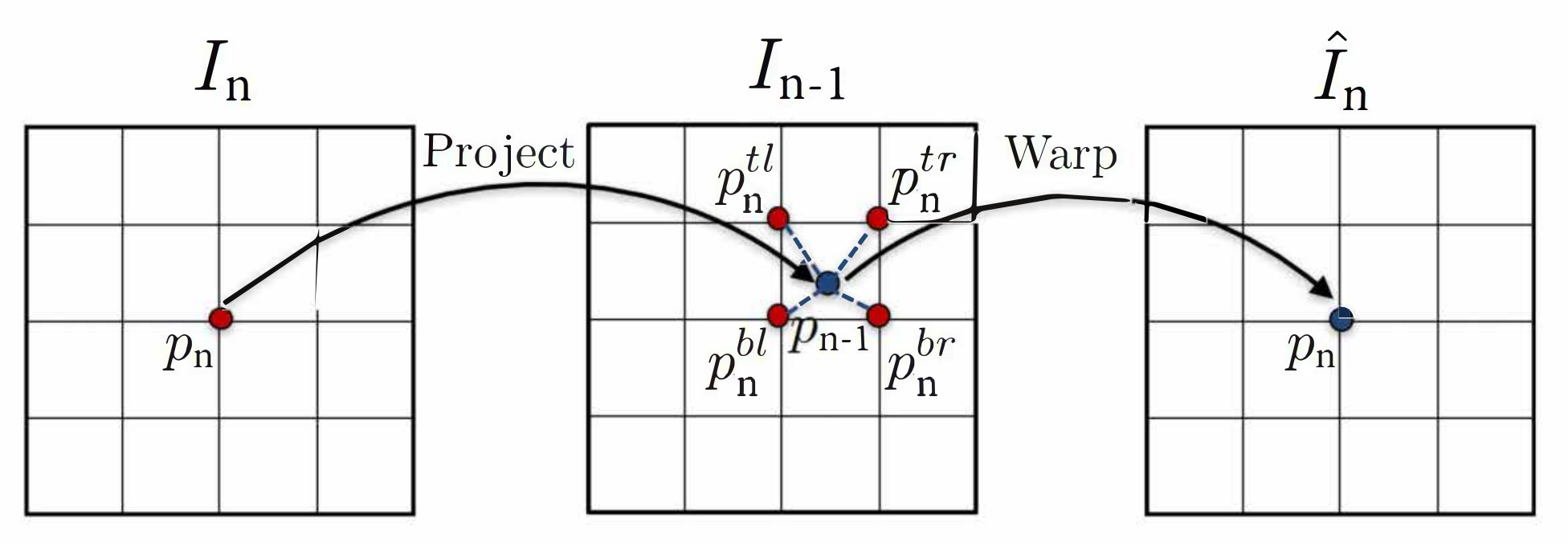}
		%\caption{fig1}
	}
	\subfigure[An illustration of unsupervised monocular depth.]{
		\includegraphics[width = 0.8\columnwidth]{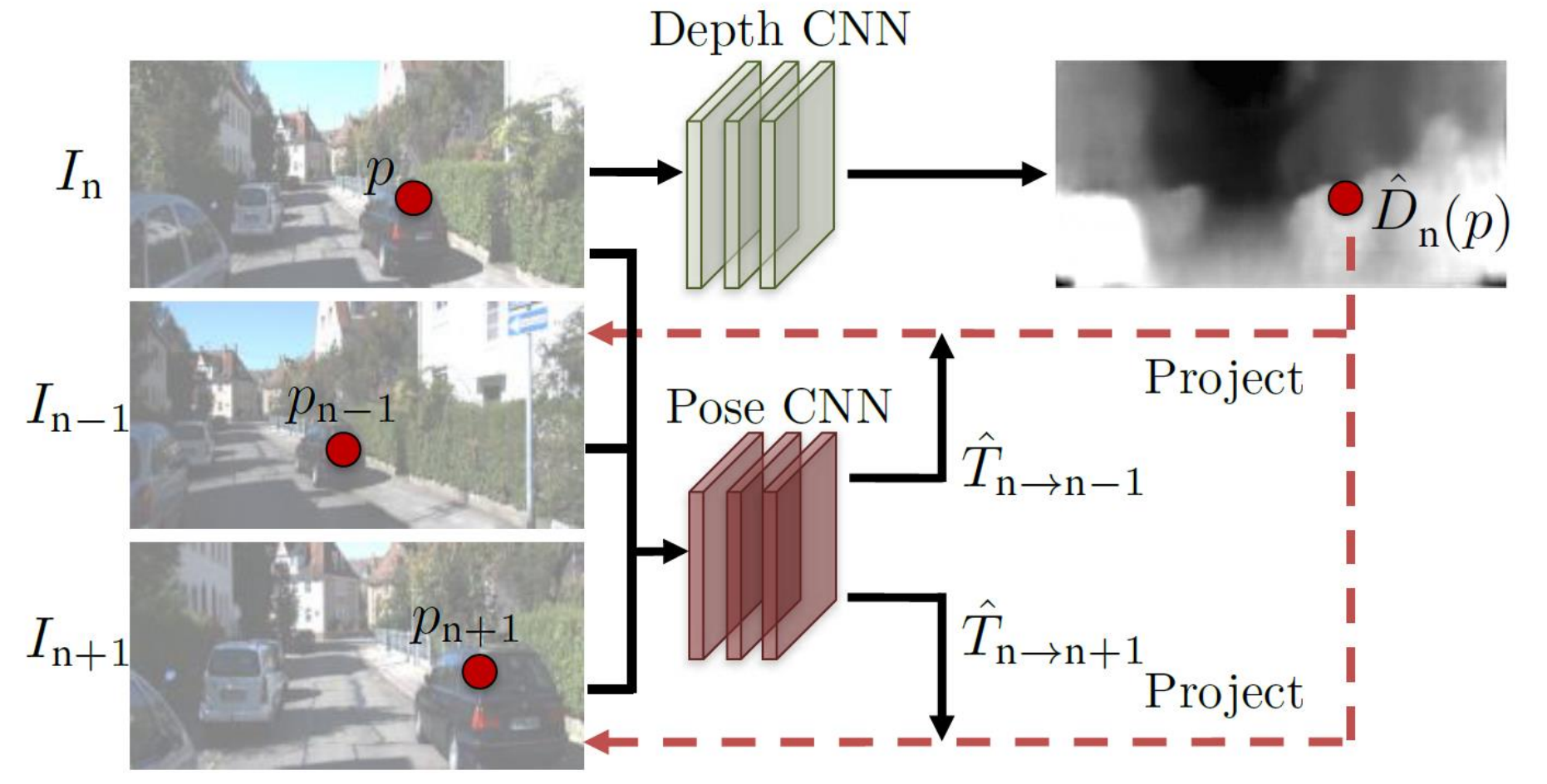}
	}
	
	\caption{Left: the image warping process for view reconstruction in unsupervised methods \cite{zhou2017unsupervised,yin2018geonet,bian2019unsupervised}.
		Right: the general framework of unsupervised monocular methods. During training, the depth $D_{n}$ and pose $\hat{T}_{n \to n-1}$ predicted by depth network and pose network are used to establish the projection relationship between $I_{n}$ and $I_{n-1}$, and then $\hat{I}_{n}$ is reconstructed by the image warping process based on projection. The differences between real $I_{n}$ and reconstructed $\hat{I}_{n}$  images are calculated to supervised the training of networks. }
	\label{fig:fig2}
\end{figure*}

\textbf{A basic model for unsupervised methods::} The unsupervised methods are trained by monocular image sequences, and the geometric constraints are built on the projection between neighboring frames:
\begin{equation}
p_{n-1} \sim K T_{n \to n-1} D_{n}(p_{n})K^{-1}p_{n}, \label{eq:10}
\end{equation}
where $p_{n}$ stands for the pixel on image $I_{n}$, and $p_{n-1}$ refers to the corresponding pixel of $p_{n}$ on image $I_{n-1}$. $K$ is the camera intrinsics matrix, which is known. $D_{n}(p_{n})$ denotes the depth value at pixel $p_{n}$, and $T_{n \to n-1}$ represents the spatial transformation between $I_{n}$ and $I_{n-1}$. Hence, if $D_{n}(p_{n})$ and $T_{n \to n-1}$ are known, the correspondence between the pixels on different images ($I_{n}$ and $I_{n-1}$) are established by projection function.
Inspired by this constraint, Zhou \textit{et al.} \cite{zhou2017unsupervised} design a depth network to predict the depth map $\hat{D}_{n}$ from a single image $I_{n}$, and a pose network to regress the transformation $\hat{T}_{n \to n-1}$ between frames ($I_{n}$ and $I_{n-1}$). Based on the output of networks, the pixel correspondences between $I_{n}$ and $I_{n-1}$ are built up:
\begin{equation}
p_{n-1} \sim K \hat{T}_{n \to n-1} \hat{D}_{n}(p_{n})K^{-1}p_{n}. \label{eq:11}
\end{equation}
Then, the photometric error between the corresponding pixels is calculated as the geometric constraints.
Zhou \textit{et al.} are inspired by \cite{szeliski1999prediction} to use a view synthesis as a metric, and the reconstruction loss is formulated as:
\begin{equation}
\mathcal{L}_{vs} = \frac{1}{N} \sum_{p}^{N}|I_{n}(p)-\hat{I}_{n}(p)|, \label{eq:12}
\end{equation}
where $p$ indexes over pixel coordinates. $\hat{I}_{n}(p)$ denotes the reconstructed frame.
The structure similarity based on SSIM is also introduced into $\mathcal{L}_{vs}$ to quantify the differences between reconstructed and target images:
\begin{equation}
\mathcal{L}_{vs} =\alpha\dfrac{1-SSIM(I_{n}-\hat{I}_{n})}{2}+ (1-\alpha)|I_{n}-\hat{I}_{n}|, \label{eq:18}
\end{equation}
where $\alpha$ is a balance factor.
Besides, the recent work \cite{godard2019digging} has proven that it is more efficient to calculate the minimum value of the reconstruction error than the mean, which has been applied in \cite{casser2019depth,bozorgtabar2019syndemo}.
The view reconstruction algorithm is applied to reconstruct the frame $\hat{I}_{n}(p)$ from $I_{n-1}$ based on the projection function, as shown in Fig. \ref{fig:fig2}. An edge-aware depth smoothness loss similar to \cite{heise2013pm,godard2017unsupervised} is adopted to encourage the local smooth of depth map:
\begin{equation}
\mathcal{L}_{smooth} = \frac{1}{N} \sum_{p}^{N} |\nabla{D}(p)| \cdot (e^{-|\nabla{I}(p)|})^{T}, \label{eq:13}
\end{equation}
Although the depth network is coupled with pose network during training, as shown in Fig. \ref{fig:fig2}, they can be used independently during testing. The above formulas (\ref{eq:11})- (\ref{eq:13}) form the basic framework of the unsupervised methods.

\textbf{Methods based on explainability mask:}
The view reconstruction algorithm based on projection function relies on the static scenario assumption, i.e., the position of dynamic objects on neighboring frames does not satisfy the projection function, which affects the photometric error and training process. Therefore, masks is widely used to reduce the influence of dynamic objects and occlusions on view reconstruction loss $\mathcal{L}_{vs}$ ( Eq. (\ref{eq:12})).
In \cite{zhou2017unsupervised} and \cite{vijayanarasimhan2017sfm}, a mask network is designed to reduce the effects of dynamic objects and occlusions on view reconstruction through:
\begin{equation}
\mathcal{L}^{M}_{vs} = \frac{1}{N} \sum_{p}^{N} M |I_{n}(p)-\hat{I}_{n}(p)|, \label{eq:17}
\end{equation}
where $M$ refers to the explainability mask predicted by a mask network. Since there is no direct supervision for $M$, training with the above loss $\mathcal{L}^{M}_{vs}$ would result in a trivial solution of the network predicting $M$ to be zero, which perfectly minimizes the loss.
Therefore, a regularization term $\mathcal{L}_{reg} (M)$ is used to encourages nonzero predictions by minimizing the cross-entropy loss with constant label 1 at each pixel location.
Besides, Vijayanarasimhan \textit{et al.} \cite{vijayanarasimhan2017sfm} design an object mask network to estimate the dynamic objects. The difference from \cite{zhou2017unsupervised} is that the object motion is regressed together with the camera pose and used to calculate the optical flow. Based on \cite{zhou2017unsupervised}, Yang \textit{et al.} \cite{yang2017unsupervised} introduce a surface normal and a depth-normal consistency term for the unsupervised framework to enhance the constraints on depth estimation. The mutual conversion between depth and normal is solved by designing a depth-to-normal layer and a normal-to-depth layer in the depth network. As a result, the depth network achieves higher accuracy than \cite{zhou2017unsupervised}.
In \cite{mahjourian2018unsupervised}, Mahjourian \textit{et al.} explore the geometric constraints between the depth map of consecutive frames. They propose an ICP loss term to enforce consistency of the estimated depth maps, and their total network framework (including mask network, pose network and depth network) are similar to \cite{zhou2017unsupervised}.

Although the mask estimation based on deep neural network is widely used in previous works \cite{zhou2017unsupervised, vijayanarasimhan2017sfm,yang2017unsupervised,mahjourian2018unsupervised} and effectively reduces the effects of dynamic objects and occlusion on reconstruction errors, it not only increases the amount of computations, but also complicates network training.
Therefore, in \cite{wang2019unsupervised} and \cite{bian2019unsupervised}, the geometry-based masks are designed to replace the masks based on deep learning and have a better effect on depth estimation.
Sun \textit{et al.} \cite{sun2019cycle} propose a cycle-consistent loss term to make full use of the sequence information.
In \cite{wang2019unsupervised}, Wang \textit{et al.} carefully consider the blank regions on the reconstructed images caused by view changes and the occlusion of the pixels generated during projection. They analyze the view reconstruction process and the influence of pixel mismatch on training. Hence, two masks on the projected image
and the target image, called the overlap mask and the blank mask, are proposed to solve the considered problems. Besides, a more detailed mask is designed to filter the trace mismatched pixels, and experiments prove the effectiveness of the proposed masks. The mask proposed by Bian \textit{et al.} \cite{bian2019unsupervised} is also based on geometry consistency constraint. They design a self-discovered mask based on the inconsistency between the depth maps of adjacent images.
Besides, a scale consistency loss term is proposed in \cite{bian2019unsupervised}, and it significantly tackles the problem of scale inconsistency between different depth maps.

\textbf{Methods based on traditional visual odometry:} Instead of using the pose estimated by a pose network, the pose regressed from traditional direct visual odometry is used to assist the depth estimation in \cite{wang2018learning}. The direct visual odometry takes the depth map generated by the depth network and a three-frame snippet to estimate the poses between frames by minimizing the photometric error; then, the calculated poses are sent back to the training framework. Therefore, because the depth network is supervised by more accurate poses, the accuracy of depth estimation is significantly improved.

\textbf{Methods based on multi-tasks framework:}
Recent approaches introduce additional networks for multi-task into the basic framework, like optical flow \cite{yin2018geonet,zou2018df}, object motion \cite{casser2019depth,ranjan2019competitive} and camera intrinsics matrix \cite{chen2019self,gordon2019depth}. Hence, the geometric relationship between different tasks is used as an additional supervision signal, which strengthens the training of the entire framework. Yin \textit{et al.} \cite{yin2018geonet} propose a jointly learning framework for depth, ego-motion and optical flow tasks. The proposed unsupervised framework consists of two parts: the rigid structure reconstructor for rigid scene reconstruction, and the non-rigid motion localizer for dynamic objects processing. A ResFlowNet is designed in the second part to learn the residual non-rigid flow. Therefore, the accuracy of all three tasks has been improved by separating rigid and non-rigid scenes and eliminating outliers through the proposed adaptive geometric consistency loss. Since the flow field of rigid regions in \cite{vijayanarasimhan2017sfm,yin2018geonet} is generated by the depth and pose estimation, errors produced by depth or pose estimation are propagated to the flow prediction. Therefore, Zou \textit{et al.} \cite{zou2018df} design an additional network to estimate the optical flow. Besides, they propose a cross-task consistency loss to constrain the consistency between the estimated flow (from network) and the generated flow (from depth and pose estimation). Ranjan \textit{et al.} \cite{ranjan2019competitive} further extend the multi-task framework, and the motion segmentation is jointly trained with other tasks (depth, pose, flow) in an unsupervised way. More tasks make the training process more complicated, so they introduce competitive collaboration to
coordinate the training process and achieve outstanding performance. Similar to \cite{vijayanarasimhan2017sfm}, Casser \textit{et al.} \cite{casser2019depth} also carefully consider the motions of dynamic objects in the scenes. An object motion network is introduced to predict the motions of individual objects, and this network takes the segmented images as input. Since the above methods are based on the prerequisites of known camera intrinsic parameters, this limits the application of the network to unknown cameras. Therefore, in \cite{chen2019self} and \cite{gordon2019depth}, they extend the pose network to estimate the camera intrinsic parameter and further reduce the prerequisites during training.

\textbf{Methods based on adversarial learning:} The adversarial learning framework is also introduced to unsupervised monocular depth estimation. Since there are no real depth maps in the unsupervised training, it is not feasible to utilize adversarial learning like Eq. (\ref{eq:12}). Therefore, instead of using the discriminator to distinguish the real and predicted depth maps, the images synthesized by view reconstruction algorithms and the real images are regarded as the input of discriminator.
In \cite{cs2018monocular,li2019sequential,almalioglu2019ganvo}, the generator consists of a pose network and a depth network, and the output of networks is used to synthesize images by view reconstruction. Then, a discriminator is designed to distinguish the real and predicted depth maps. Since temporal information helps to improve the performance of the network, the LSTM module is introduced to the pose network and depth network to contact contextual information in \cite{li2019sequential,almalioglu2019ganvo}. Furthermore, Li \textit{et al.} \cite{li2019sequential} design an additional network to eliminate the shortcoming of view reconstruction algorithm, which is similar to \cite{zhou2017unsupervised}. In order to get more 3D cues, the information between frames extracted by LSTM and the single images are adopted together to depth estimation.

Compared with supervised and semi-supervised methods, unsupervised methods learn the depth information from the geometric constraints instead of ground truth. Therefore, the training process relies on the monocular sequences captured by a camera, and unsupervised learning is beneficial for the practical application of unsupervised methods. However, because of learning from monocular sequences, which do not contain the absolute scale information, unsupervised methods suffer from scale ambiguity, scale inconsistency, occlusions and other problems.

\subsection{Semi-supervised monocular depth estimation}

Since there is no need for ground truth during training, the performance of unsupervised methods is still far from the supervised methods. Besides, the unsupervised methods also suffer from various problems, like scale ambiguity and scale inconsistency. Therefore, the semi-supervised methods are proposed to get higher estimation accuracy while reducing the dependence on the expensive ground truth. Besides, the scale information can be learned from the semi-supervised signals.

Training on stereo image pairs is similar to the case of monocular videos, and the main difference is whether the
transformation between two frames (left-right images or front-back images) is known. Therefore, some studies regard the framework based on stereo image pairs as unsupervised methods \cite{garg2016unsupervised}, while others treat them as semi-supervised methods \cite{zhu2017unpaired}. In this review, we consider them the semi-supervised methods, and the poses between left-right images are the supervised signals during training.

\begin{figure}[t]
	\centering
	\includegraphics[width = \columnwidth]{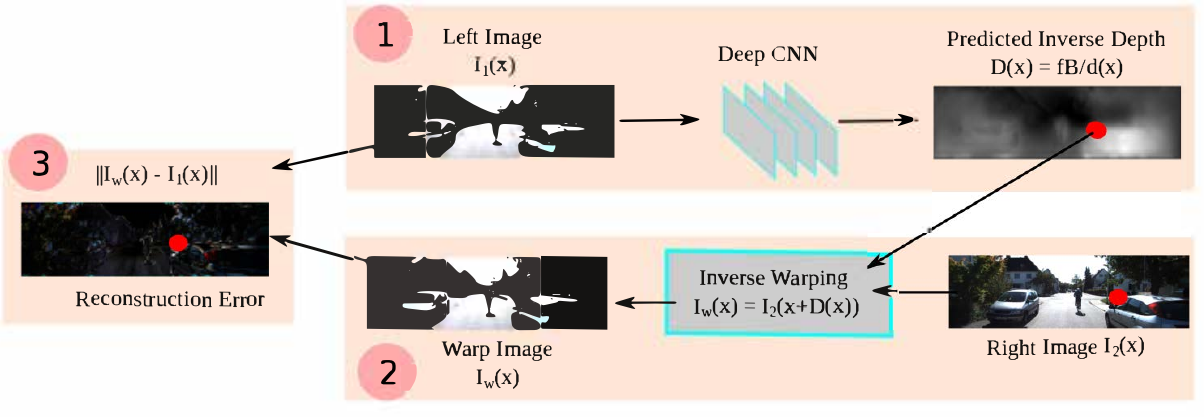}
	%\caption{fig1}
	\caption{The general framework of semi-supervised monocular depth estimation based on stereo image pairs, which is proposed in \cite{garg2016unsupervised}. The depth network takes the left image to predict its pixel-level inverse depth map (or disparity map), and the predicted inverse depth map is used to reconstruct the left image from right image by inverse warping algorithm. The reconstruction error is calculated to supervised the training process.}
	\label{fig:fig4}
\end{figure}

\textbf{A basic model for semi-supervised methods::} Semi-supervised methods trained on stereo image pairs estimate the disparity maps (inverse depth maps) between the left and right images. Then, the disparity map $Dis$ calculated from predicted inverse depth is used to synthesize the left image from the right image by inverse warping, as shown in Fig. \ref{fig:fig4}. Similar to the unsupervised methods, the differences between the synthesized images $I_{w}$ and real images $I_{l}$ are used as a supervised signal and to constrain the training process:
\begin{equation}
\begin{aligned}
\mathcal{L}_{recons} & =  \sum_{p}||I_{l}(p)-I_{w}(p)||^{2}, \\
&= \sum_{p}||I_{l}(p)-I_{r}(p+Dis(p))||^{2},
\end{aligned} \label{eq:14}
\end{equation}
where $I_{r}$ is the corresponding right images. The depth map $d$ can be transferred from the predicted disparity map through: $d = fB/D$, where $f$ is the local length of cameras, and $B$ refers to the distance between left and right cameras.
Based on the above framework, Garg \textit{et al.} \cite{garg2016unsupervised} also use a smoothness loss term to improve the continuities of disparity maps.
Godard \textit{et al.} \cite{godard2017unsupervised} improve both the above network framework and the loss functions.
The right disparity map $Dis^{r}$ is predicted together with the left disparity map $Dis^{l}$ and used to reconstruct the right image from left image. Besides, they present a left-right disparity consistency loss to constrain the consistency between left and right disparities:
\begin{equation}
\mathcal{L}_{lr} =  \frac{1}{N} \sum_{p}|Dis^{l}(p)-Dis^{r}(p+Dis^{l}(p))|, \label{eq:15}
\end{equation}
Besides, the SSIM \cite{wang2004image} is introduced to strengthen the structure similarity between the synthesised images and real images, and the loss function is similar to Eq.\ref{eq:18}.
As a result, the experiments demonstrate the effectiveness of these improvements, and the performance outperforms the previous works \cite{garg2016unsupervised}. Considering that above framework suffers from occlusions and left image border, a framework based on trinocular assumptions is proposed in \cite{poggi2018learning}. In \cite{ramirez2018geometry}, Ramirez \textit{et al.} propose a framework for joint depth and semantic prediction tasks. An additional decoder stream is designed to estimate semantic labels and trained in a supervised way. Furthermore, a cross-domain discontinuity term based on the predicted semantic image is applied to improve the smoothness of the predicted depth map, which shows a better performance than the previous smoothness loss terms (like Eq. (\ref{eq:11})). Similarly, Chen \textit{et al.} \cite{chen2019towards} also leverage semantic segmentation to improve the monocular depth estimation. In \cite{chen2019towards}, depth estimation and semantic segmentation share the same network framework, and switch by condition. A novel left-right semantic consistency term is proposed to perform region-aware depth estimation and improves the accuracy and robustness of both tasks.

\textbf{Methods based on stereo matching:}
Luo \textit{et al.} \cite{luo2018single} present a view synthesis network based on Deep3D \cite{xie2016deep3d} to estimate the right image from the left image, which is different from above works. Moreover, a stereo matching network is designed to take the raw left and synthesised right images to regress the disparity map. During training, the view synthesis network is supervised by the raw right images to improve the construction quality, and the predicted disparity maps are used to reconstruct the left images from the estimated right images. Similar to \cite{luo2018single}, Tosi \textit{et al.} \cite{tosi2019learning} also leverage the stereo matching strategy to improve the performance and robustness of monocular depth estimation. Features from different viewpoints are synthesised by performing stereo matching, thereby achieving outstanding performance. Their network framework consists of three parts: a multi-scale feature extractor for high-level feature extraction, a disparity network for disparity map prediction, and a refinement network for disparity refinement. In comparison to \cite{luo2018single}, the networks proposed in \cite{tosi2019learning} are jointly trained while those in \cite{luo2018single} are trained independently, thereby simplifying the complexity of training in \cite{tosi2019learning}.

\textbf{Methods based on adversarial learning and knowledge distillation:}
Combining advanced network frameworks, like adversarial learning \cite{aleotti2018generative,pilzer2018unsupervised,feng2019sganvo} and knowledge distillation \cite{pilzer2019refine}, is becoming popular and can significantly improve the performance.
The framework of knowledge distillation consists of two neural networks, a teacher network and a student network. Teacher network is more complex than the student network. The purpose of knowledge distillation is to transfer the knowledge learned by the teacher network to the student network, so that the functions learned by the large model are compressed into smaller and faster models. Pilzer \textit{et al.} \cite{pilzer2019refine} follow this idea, and knowledge distillation is used to transfer information from the refinement network to the student network. Considering the effectiveness of training with synthetic images, Zhao \textit{et al.} \cite{zhao2019geometry} adopt the framework of cycle GAN for the transformation between the synthetic and real domains to expand the data set, and propose a geometry-aware symmetric domain adaptation network (GASDA) to make better use of the synthetic data. Their network learns from the ground truth labels in synthetic domain as well as the epipolar geometry of the real domain, thereby achieving competitive results.  Wu \textit{et al.} \cite{wu2019spatial} improve the architecture of the generator by utilizing a spatial correspondence module for feature matching and an attention mechanism for feature re-weighting.

\textbf{Methods based on sparse ground truth:} To strengthen the supervised signals, the sparse ground truth is widely incorporated into the training framework.
Kuznietsov \textit{et al.} \cite{kuznietsov2017semi} adopt the ground truth depth collected by LIDAR for semi-supervised learning. Besides, both the left and right depth maps ($D_{l},D_{r}$) are estimated by CNNs, and the supervision signal based on LIDAR data ($G_{l},G_{r}$) is formulated as:
\begin{equation}
\begin{aligned}
\mathcal{L}_{recons} &=  \sum_{p \in \Omega_{Z,l}}||D_{l}(p)-G_{l}(p)||_{\delta}, \\
&= \sum_{p \in \Omega_{Z,r}}||D_{r}(p)-G_{r}(p)||_{\delta},
\end{aligned} \label{eq:16}
\end{equation}
where $\Omega_{Z,l}$ refers to the set of pixels with available ground truth, and $||*||_{\delta}$ denotes the berHu norm \cite{zwald2012berhu}.
Similarly, based on \cite{godard2017unsupervised}, He \textit{et al.} \cite{he2018wearable} introduce the loss between the predicted depth maps and LIDAR data as an additional signal. Moreover, the physical information is also adopted into the semi-supervised methods. Fei \textit{et al.} \cite{fei2019geo} use the global orientation computed from inertial measurements as a priori information to constrain the normal vectors to surfaces of objects. Generally, normal vectors to surfaces of objects are parallel or perpendicular to the direction of gravity, and they can easily calculate from the estimated depth map. Therefore, this physical priori significantly improves the accuracy of depth estimation.

Semi-supervised methods achieve a better accuracy than unsupervised methods because of the semi-supervised signals, and the scale information can be learned from these signals. However, the accuracy of semi-supervised methods relies heavily on the ground truth, like pose and LIDAR data, although they are easier to get than expensive dense depth maps.

%\subsection{Summary}

%Because of being supervised by the ground truth, the supervised methods can effectively learn the 3D structures and their scale information from single images. They outperform the semi-supervised and unsupervised methods by learning from ground truth directly. However, these supervised methods are limited by the labeled training sets, which are hard and expensive to acquire.
%Semi-supervised methods achieve a better accuracy than unsupervised methods by the semi-supervised signals, and the scale information can be learned from these signals. However, the accuracy of semi-supervised methods relies heavily on the ground truth, like pose and LIDAR data, although they are easier to get than expensive dense depth maps.
%Compared with supervised and semi-supervised methods, unsupervised methods learn the depth information from the geometric constraints instead of ground truth. Therefore, the training process does not require any ground truth but monocular sequences captured by a camera, and it is beneficial for the practical application of unsupervised methods. Because of learning from monocular sequences, which do not contain the scale information, unsupervised methods suffer from scale ambiguity, scale inconsistency and other problems.

\subsection{Applications}

The monocular depth estimation based on deep learning has been widely applied in SLAM (or visual odometry (VO)) to improve the mapping process, recover the absolute scale, and replace the RGB-D sensor in dense mapping. \textit{Improving the map:} LOO \textit{et al.} \cite{loo2019cnn} introduce the monocular depth prediction into the SVO framework \cite{forster2014svo}, and the depth value predicted by deep neural networks is used to initialize the mean and variance of the depth at a feature location. Therefore, the depth uncertainty during mapping is effectively reduced with the help of introducing depth prediction, thereby improving the map built by CNN-SVO. \textit{Scale recovery:} Since the depth neural network can predict the depth containing absolute scale information from a single image, the scale ambiguity and scale drift of monocular VO methods \cite{engel2017direct} can be effectively solved with the help of deep learning-based depth estimation. Yin \textit{et al.} \cite{yin2017scale} and Yang \textit{et al.} \cite{yang2018deep} follow this idea and leverage the depth estimation based on deep learning to recover the absolute scale of monocular VO. \textit{Replace the RGB-D sensor:} As reviewed by \cite{tang2020perception}, most of dense SLAM methods take RGB-D sensor to build the dense maps of scenes. Compared with RGB-D sensors, depth networks can generate the accurate and dense depth maps from the single images captured by monocular cameras. Besides, monocular cameras have the advantages of small size, low power consumption, and easy access. Therefore, Tateno \textit{et al.} \cite{tateno2017cnn} propose a method that introduce the deep depth estimation into dense monocular reconstruction, and their methods also demonstrate the effectiveness of deep learning-based depth prediction in the absolute scale recovery.

\begin{table*}[]
	
	\scriptsize
	
	\centering
	
	\caption{A summary of deep learning-based monocular depth estimation. ``Mono.'' refers to ``Monocular'', and ``multi-tasks'' means that in addition to pose and depth estimation, there are other tasks that are jointly trained in the framework, such as semantic segmentation, motion segmentation, optical flow, etc.}
	
	\label{Tab01}
	\resizebox{2\columnwidth}{!}{
		\begin{tabular}{c|c|c|ccc|c|}
			\toprule
			\multicolumn{1}{c}{}&\multicolumn{1}{c}{}&\multicolumn{1}{c}{}&\multicolumn{3}{c}{Supervised (Sup) manner}& \multicolumn{1}{c}{} \\
			\cmidrule(r){4-6}
			\hline
			Methods  &Years &Training set &Sup &Semi-sup &Unsup  &Main contributions  \\
			\hline
			Eigen \textit{et al.} \cite{eigen2014depth} & 2014 & RGB + Depth &  $\surd$ & &  & CNNs \\
			Li \textit{et al.} \cite{li2015depth}       & 2015 & RGB + Depth &  $\surd$ & &  & hierarchical CRFs\\
			Liu \textit{et al.} \cite{liu2015learning}  & 2015 & RGB + Depth &  $\surd$ & &  & continuous CRF\\
			Wang \textit{et al.} \cite{wang2015towards} & 2015 & RGB + Depth &  $\surd$ & &  & Semantic labels, hierarchical CRFs\\
			Shelhamer \textit{et al.}  \cite{shelhamer2015scene} & 2015 & RGB + Depth     &  $\surd$ & &  & Fully CNNs\\
			Eigen \textit{et al.}  \cite{eigen2015predicting}    & 2015 & RGB + Depth     &  $\surd$ & &  & Multi-task\\
			Szegedy \textit{et al.}  \cite{szegedy2015going}     & 2015 & RGB + Depth     &  $\surd$ & &  & Inception Module\\
			Mousavian \textit{et al.} \cite{mousavian2016joint}  & 2016 & RGB + Depth     &  $\surd$ & &  &Multi-task \\
			Roy \textit{et al.}  \cite{roy2016monocular}         & 2016 & RGB + Depth     &  $\surd$ & &  & RFs\\
			Mayer \textit{et al.} \cite{mayer2016large}          & 2016 & RGB + Disparity &  $\surd$ & &  & Multi-task\\
			Laina \textit{et al.} \cite{laina2016deeper}         & 2016 & RGB + Depth     &  $\surd$ & &  & Residual learning\\
			Jung \textit{et al.}  \cite{jung2017depth}           & 2017 & RGB + Depth     &  $\surd$ & &  & Adversarial learning\\
			Kendall \textit{et al.} \cite{kendall2017end}        & 2017 & Stereo images + Disparity &  $\surd$ & &  & Disparity Loss\\
			Zhang \textit{et al.} \cite{zhang2018joint}          & 2018 & RGB + Depth &  $\surd$ & &  & Task-attentional, BerHu loss\\
			Xu \textit{et al.} \cite{xu2018structured}           & 2018 & RGB + Depth &  $\surd$ & &  & Continuous CRF, structured attention\\
			%Zhang \textit{et al.} \cite{zhang2018joint}          & 2018 & RGB + Depth &  $\surd$ & &  & BerHu Loss\\
			Gwn \textit{et al.}  \cite{gwn2018generative}        & 2018 & RGB + Depth &  $\surd$ & &  & Conditional GAN\\
			Fu \textit{et al.} \cite{fu2018deep}                 & 2018 & RGB + Depth &  $\surd$ & &  & Ordinal regression\\
			Facil \textit{et al.} \cite{facil2019cam}            & 2019 & RGB + Depth &  $\surd$ & &  & Camera model\\
			Woft \textit{et al.} \cite{wofk2019fastdepth}        & 2019 & RGB + Depth &  $\surd$ & &  & Lightweight network\\
			\hline
			Garg \textit{et al.} \cite{garg2016unsupervised}     & 2016 & Stereo images   & &  $\surd$  & & Stereo framework \\
			Chen \textit{et al.} \cite{chen2016single}           & 2016 & RGB + Relative depth annotations & &  $\surd$  & & The wild scene\\
			Godard \textit{et al.} \cite{godard2017unsupervised} & 2017 & Stereo images   & &  $\surd$  & & Left-right consistency loss \\
			Kuznietsov \textit{et al.} \cite{kuznietsov2017semi} & 2017 & Stereo images + LiDAR  & &  $\surd$  &   &Direct image alignment loss \\
			Poggi \textit{et al.} \cite{poggi2018learning}       & 2018 & Stereo images   & &  $\surd$  &   & Trinocular assumption \\
			Ramirez \textit{et al.} \cite{ramirez2018geometry}   & 2018 & Stereo images + Semantic Label  & &  $\surd$  &   & Semantic prediction\\
			Aleotti \textit{et al.} \cite{aleotti2018generative} & 2018 & Stereo images   & &  $\surd$  &   &   GAN\\
			Pilzer \textit{et al.} \cite{pilzer2018unsupervised} & 2018 & Stereo images   & &  $\surd$  &   &  Cycled generative network\\
			Luo \textit{et al.} \cite{luo2018single}             & 2018 & Stereo images   & &  $\surd$ &  & Stereo matching, view prediction\\
			He \textit{et al.} \cite{he2018wearable}             & 2018 & Stereo images +LIDAR      & &  $\surd$ &  & Weak-supervised framework\\
			Pilzer \textit{et al.} \cite{pilzer2019refine}       & 2019 & Stereo images   & &  $\surd$  &   &  Knowledge distillation\\
			Tosi \textit{et al.} \cite{tosi2019learning}         & 2019 & Stereo images   & &  $\surd$  &   & Stereo matching \\
			Chen \textit{et al.} \cite{chen2019towards}          & 2019 & Stereo images   & &  $\surd$  &   & Multi-task\\
			Fei \textit{et al.} \cite{fei2019geo}                & 2019 & Stereo images + IMU + Semantic Label  &   &  $\surd$  &   & Multi-task,physical information \\
			Feng \textit{et al.}   \cite{feng2019sganvo}         & 2019 & Stereo images   & &  $\surd$ &  & Stacked-GAN\\
			Wang \textit{et al.} \cite{wang2018learning}         & 2018 & Mono. sequences & &  $\surd$  &  & Direct VO\\
			Zhan \textit{et al.} \cite{zhan2018unsupervised}     & 2018 & Stereo sequences& &  $\surd$  &  & Deep feature reconstruction\\
			Li \textit{et al.} \cite{li2018undeepvo}             & 2018 & Stereo sequences& &  $\surd$ &  & Absolute  scale recovery\\
			Wang \textit{et al.} \cite{wang2019unos}             & 2019 & Stereo sequences& &  $\surd$ &  & Multi-task\\
			Zhao \textit{et al.} \cite{zhao2019geometry}         & 2019 & Stereo images + Synthesized GT  & &  $\surd$ &  & Domain adaptation, cycle GAN\\
			Wu \textit{et al.} \cite{wu2019spatial}              & 2019 & Mono. sequences + LIDAR         & &  $\surd$ &  & Attention mechanism, GAN\\
			\hline
			Zhou \textit{et al.} \cite{zhou2017unsupervised}     & 2017 & Mono. sequences & &   & $\surd$ & Monocular framework, mask network \\
			Vijayanarasimhan \textit{et al.} \cite{vijayanarasimhan2017sfm}& 2017 & Mono. sequences &  &   & $\surd$ & Multi-task\\
			Yang \textit{et al.} \cite{yang2017unsupervised}     & 2017 & Mono. sequences &  &   & $\surd$ & Surface normal\\
			Mahjourian \textit{et al.} \cite{mahjourian2018unsupervised}   & 2018 & Mono. sequences &  &   & $\surd$ & ICP loss\\
			Yin \textit{et al.} \cite{yin2018geonet}             & 2018 & Mono. sequences &  &   & $\surd$ & Multi-task\\
			Zou \textit{et al.} \cite{zou2018df}                 & 2018 & Mono. sequences &  &   & $\surd$ & Multi-task \\
			Kumar \textit{et al.} \cite{cs2018monocular}         & 2018 & Mono. sequences &  &   & $\surd$ &GAN \\
			Sun \textit{et al.} \cite{sun2019cycle}              & 2019 & Mono. sequences &  &   & $\surd$ & Cycle-consistent loss\\
			Wang \textit{et al.} \cite{wang2019unsupervised}     & 2019 & Mono. sequences &  &   & $\surd$ & Geometry mask\\
			Bian \textit{et al.} \cite{bian2019unsupervised}     & 2019 & Mono. sequences &  &   & $\surd$ & Scale-consistency\\
			Casser \textit{et al.} \cite{casser2019depth}        & 2019 & Mono. sequences &  &   & $\surd$ & Object motion prediction\\
			Ranjan \textit{et al.} \cite{ranjan2019competitive}  & 2019 & Mono. sequences &  &   & $\surd$ & Multi-task\\
			Chen \textit{et al.} \cite{chen2019self}             & 2019 & Mono. sequences &  &   & $\surd$ & Camera intrinsic prediction\\
			Gordon \textit{et al.} \cite{gordon2019depth}        & 2019 & Mono. sequences &  &   & $\surd$ & Camera intrinsic prediction\\
			Li \textit{et al.} \cite{li2019sequential}           & 2019 & Mono. sequences &  &   & $\surd$ & GAN, LSTM, mask\\
			Almalioglu \textit{et al.} \cite{almalioglu2019ganvo}& 2019 & Mono. sequences &  &   & $\surd$ & GAN, LSTM\\
			
			\bottomrule
			
	\end{tabular}}
	
\end{table*}

\begin{table*}[]
	
	\scriptsize
	
	\centering
	
	\caption{ Monocular depth results of semi-supervised and unsupervised methods on the KITTI dataset\cite{geiger2012we}. ``Cap'' stands for the upper limit of predicted depth, and ``sup'' refers to ``supervised ''. We show the best results in bold. Note that this table does not contain the results with ``online refinement'' or ``pretrain'' (``CS+K'' ``ImageNet pretrain'').}
	
	\label{Tab02}
	
	\begin{tabular}{c|c|c|c|cccc|ccc}
		
		\toprule
		\multicolumn{4}{c}{}& \multicolumn{4}{c}{Lower is better} & \multicolumn{3}{c}{Accuracy: higher is better}  \\
		\cmidrule(r){1-4}\cmidrule(r){5-8} \cmidrule(r){9-11}
		\hline
		Method   &   Year 	&  Training pattern    &   Cap		&  Abs Rel      &  Sq Rel    &   RMSE 	&  RMSE log     &   $\delta < 1.25^{1}$		&  $\delta < 1.25^{2}$      &  $\delta < 1.25^{3}$ \\
		\hline
		\hline
		Garg \textit{et al.} \cite{garg2016unsupervised}L12 Aug8 $\times$ cap 50m& 2016 & Semi-sup & 50m & 0.169 & 1.080 & 5.104 & 0.273 & 0.740 & 0.904 & 0.962 \\
		Godard \textit{et al.} \cite{godard2017unsupervised} & 2017 & Semi-sup & 80m & 0.148 & 1.344 & 5.927 & 0.247 & 0.803 & 0.922 & 0.964 \\
		Kuznietsov \textit{et al.} \cite{kuznietsov2017semi} & 2017 & Semi-sup & 80m & 0.113 & 0.741 & \textbf{4.621} & \textbf{0.189} & 0.803 & \textbf{0.960} & \textbf{0.986} \\
		Poggi \textit{et al.} \cite{poggi2018learning}       & 2018 & Semi-sup & 80m & 0.126 & 0.961 & 5.205 & 0.220 & 0.835 & 0.941 & 0.974\\
		Ramirez \textit{et al.} \cite{ramirez2018geometry}   & 2018 & Semi-sup & 80m & 0.143 & 2.161 & 6.526 & 0.222 & 0.850 & 0.939 & 0.972\\
		Aleotti \textit{et al.} \cite{aleotti2018generative} & 2018 & Semi-sup & 80m & 0.119 & 1.239 & 5.998 & 0.212 & 0.846 & 0.940 & 0.976\\
		Pilzer \textit{et al.} \cite{pilzer2018unsupervised} & 2018 & Semi-sup & 80m & 0.152 & 1.388 & 6.016 & 0.247 & 0.789 & 0.918 & 0.965\\
		Luo \textit{et al.} \cite{luo2018single}             & 2018 & Semi-sup & 80m & 0.102 & \textbf{0.700} & 4.681 & 0.200 & 0.872 & 0.954 & 0.978\\
		He \textit{et al.} \cite{he2018wearable}             & 2018 & Semi-sup & 80m & 0.110 & 1.085 & 5.628 & 0.199 & 0.855 & 0.949 & 0.981\\
		Pilzer \textit{et al.} \cite{pilzer2019refine}  & 2019 & Semi-sup & 80m & \textbf{0.098} & 0.831 & 4.656 & 0.202 & \textbf{0.882} & 0.948 & 0.973 \\
		Tosi \textit{et al.} \cite{tosi2019learning}    & 2019 & Semi-sup & 80m & 0.111 & 0.867 & 4.714 & 0.199 & 0.864 & 0.954 & 0.979 \\
		Chen \textit{et al.} \cite{chen2019towards}     & 2019 & Semi-sup & 80m & 0.118 & 0.905 & 5.096 & 0.211 & 0.839 & 0.945 & 0.977 \\
		Godard \textit{et al.} \cite{godard2019digging} & 2019 & Semi-sup & 80m & 0.127 & 1.031 & 5.266 & 0.221 & 0.836 & 0.943 & 0.974 \\
		Watson \textit{et al.} \cite{watson2019self}    & 2019 & Semi-sup & 80m & 0.112 & 0.857 & 4.807 & 0.203 & 0.862 & 0.952 & 0.978 \\
		\hline
		\hline
		Zhou \textit{et al.} \cite{zhou2017unsupervised}      & 2017 & Unsup & 80m & 0.208 & 1.768 & 6.865 & 0.283 & 0.678 &  0.885 & 0.957 \\
		Yang \textit{et al.} \cite{yang2017unsupervised}      & 2017 & Unsup & 80m & 0.182 & 1.481 & 6.501 & 0.267 & 0.725 & 0.906 & 0.963 \\
		Mahjourian \textit{et al.} \cite{mahjourian2018unsupervised}   & 2018 & Unsup & 80m & 0.163 & 1.240 & 6.221 & 0.250 & 0.762 & 0.916 & 0.968 \\
		Yin \textit{et al.} \cite{yin2018geonet}              & 2018 & Unsup & 80m & 0.155 & 1.296 & 5.857 & 0.233 & 0.793 & 0.931 & 0.973 \\
		Zou \textit{et al.} \cite{zou2018df}                  & 2018 & Unsup & 80m & 0.150 & 1.124 & 5.507 & 0.223 & 0.806 & 0.933 & 0.973 \\
		Wang \textit{et al.} \cite{wang2019unsupervised}      & 2019 & Unsup & 80m & 0.158 & 1.277 & 5.858 & 0.233 & 0.785 & 0.929 & 0.973 \\
		Bian \textit{et al.} \cite{bian2019unsupervised}      & 2019 & Unsup & 80m & 0.137 & 1.089 & 5.439 & 0.217 & 0.830 & 0.942 & 0.975 \\
		Casser \textit{et al.} (M) \cite{casser2019depth}     & 2019 & Unsup & 80m & 0.141 & \textbf{1.026} & 5.291 & 0.215 & 0.816 & 0.945 & 0.979 \\
		Ranjan \textit{et al.} \cite{ranjan2019competitive}   & 2019 & Unsup & 80m & 0.140 & 1.070 & 5.326 & 0.217 & 0.826 & 0.941 & 0.975 \\
		Chen \textit{et al.} \cite{chen2019self}              & 2019 & Unsup & 80m & 0.135 & 1.070 & 5.230 & \textbf{0.210} & 0.841 & \textbf{0.948} & \textbf{0.980} \\
		Li \textit{et al.} \cite{li2019sequential}            & 2019 & Unsup & 80m & 0.150 & 1.127 & 5.564 & 0.229 & 0.823 & 0.936 & 0.974 \\
		Almalioglu \textit{et al.} \cite{almalioglu2019ganvo} & 2019 & Unsup & 80m & 0.150 & 1.141 & 5.448 & 0.216 & 0.808 & 0.939 & 0.975 \\
		Godard \textit{et al.} \cite{godard2019digging}       & 2019 & Unsup & 80m & \textbf{0.132} & 1.044 & \textbf{5.142} & \textbf{0.210} & \textbf{0.845} & \textbf{0.948} & 0.977 \\
		\bottomrule
	\end{tabular}
\end{table*}

\section{Discussion}
In general, we think that the development of monocular depth estimation will still focus on improving the accuracy, transferability, and real-time performance.

\textbf{Accuracy:} Most of the previous works mainly focus on improving the accuracy of depth estimation by adopting new loss functions or network frameworks, as shown in Table \ref{Tab01}. Several well-known network frameworks, like \textit{LSTM, VAE, GANs}, have shown their effectiveness in improving the performance of depth estimation. Therefore, with the development of deep neural networks, trying new network frameworks, like \textit{3D convolution} \cite{cheng2018learning}, \textit{graph convolution} \cite{li2018deeper}, \textit{attentional mechanism} \cite{vaswani2017attention} and \textit{knowledge distillation}  \cite{ahn2019variational}, may get satisfactory results.
Although the unsupervised methods do not rely on ground truth during training, their accuracy is far from the current most effective semi-supervised methods, as shown in Table \ref{Tab02}.
Finding a \textit{more efficient geometric constraint to improve the unsupervised methods \cite{godard2019digging}} may be a good direction.
For example, \textit{the target-level dynamic object motion estimation combined with geometry-based mask}, will be an effective solution to the impact of dynamic objects and occlusions on view reconstruction.
Besides, the unsupervised methods training on monocular videos suffer from \textit{scale ambiguity} and \textit{scale inconsistency}. Although some loss terms are proposed to constrain the scale consistency, this problem is not solved well. Since the semantic information is mainly used to constrain the smoothness of depth map during training, it will be a good research direction for solving monocular scale ambiguity by \textit{learning the scale from semantic information}.
Moreover, \textit{the multi-task joint training combining with the geometric relationship between tasks} is also a proven method that is worthy of further study.
To get a state-of-the-art result, the network framework is becoming more and more complicated, and the loss terms are becoming more complicated, which make the training of the network difficult. Furthermore, the increase of loss terms will also pose a challenge on the selection of hyperparameters.
%\textit{Reducing the impact of hyperparameters on the network} or
A more effective way for \textit{designing deep learning-based hyperparameter setting methods} is also a huge challenge. For example, \textit{estimating the intrinsics matrix of the monocular camera and the parameters of stereo cameras based on deep learning} may be a promising direction.

\textbf{Transferability:} Transferability refers to the performance of the same network on different cameras, different scenarios, and different datasets. The transferability of depth networks is raising increasing attention. Most of the current methods are trained and tested on the same dataset, thereby achieving a satisfactory result. However, the training set and testing set in different domains or collected by different cameras often lead to severe performance degradation. \textit{Incorporating camera parameters into depth estimation framework and leveraging domain adaptation technology during training} will significantly improve the transferability of depth network, and they are becoming a hot topic recently.

%\textbf{Convenience:} Convenience stands for the convenience of network training, testing and practical use.

\textbf{Real-time performance:} Although deeper networks show outstanding performance, they require more computation time to complete estimation tasks, which is a great challenge for their applications. The ability of depth estimation networks to run in real-time on embedded devices will have significant implications for their practical applications. Therefore, the development of \textit{lightweight networks based on supervised, semi-supervised and unsupervised learning} will be a promising direction, and there are not much related researches in this field at present. As the number of parameters of the lightweight network is smaller, this affects the performance of the network. Therefore, it is a worthwhile subject to improve accuracy while ensuring real-time performance.

In addition, there is very little researches on \textit{the mechanism of monocular depth estimation methods based on deep learning}, like what depth networks have learned and what depth cues they exploit. For example, the research in \cite{dijk2019neural} focuses on the cues of neural network learning depth from a single image, and its experiments have shown that current depth networks ignore the apparent size of known obstacles, which is different from how humans perceive depth information. Therefore, studying the mechanism of depth estimation is a promising direction, which may effectively improve the accuracy, transferability and real-time performance. The application of monocular depth estimation in environmental perception \cite{tang2020perception} and control \cite{wu2016stability,tang2020input} of autonomous robots is also a direction worthy of research.
%whether it is a supervised, semi-supervised or unsupervised method. We believe that only with a good understanding of the mechanism can we propose effective solutions.

\section{Conclusion}
In this review, we aim to contribute to this growing area of research in deep learning-based monocular depth estimation. Therefore, we survey the related works of monocular depth estimation from the aspect of training manner, including supervised, unsupervised as well as semi-supervised learning, combining with the application of loss functions and network frameworks. In the end, we also discuss the current hot topics as well as challenges and provide some valuable ideas and promising directions for future researches.

%\section*{Acknowledgment}

%This work was supported by the National Key Research and Development Program of China under Grant 2018YFC0809302, the National Natural Science Foundation of China under Grant Nos. 61751305, 61673176, by the Programme of Introducing Talents of Discipline to Universities (the 111 Project) under Grant B17017.

%\begin{thebibliography}{1}

  %0.5em minus 0.4em\relax Harlow, England: Addison-Wesley, 1999.

%\end{thebibliography}

{%\small
	\bibliographystyle{IEEEtran}
	\bibliography{egbib}

\begin{thebibliography}{100}
\providecommand{\url}[1]{#1}
\csname url@samestyle\endcsname
\providecommand{\newblock}{\relax}
\providecommand{\bibinfo}[2]{#2}
\providecommand{\BIBentrySTDinterwordspacing}{\spaceskip=0pt\relax}
\providecommand{\BIBentryALTinterwordstretchfactor}{4}
\providecommand{\BIBentryALTinterwordspacing}{\spaceskip=\fontdimen2\font plus
\BIBentryALTinterwordstretchfactor\fontdimen3\font minus
  \fontdimen4\font\relax}
\providecommand{\BIBforeignlanguage}[2]{{%
\expandafter\ifx\csname l@#1\endcsname\relax
\typeout{** WARNING: IEEEtran.bst: No hyphenation pattern has been}%
\typeout{** loaded for the language `#1'. Using the pattern for}%
\typeout{** the default language instead.}%
\else
\language=\csname l@#1\endcsname
\fi
#2}}
\providecommand{\BIBdecl}{\relax}
\BIBdecl

\bibitem{hu2012robust}
G.~Hu, S.~Huang, L.~Zhao, A.~Alempijevic, and G.~Dissanayake, ``A robust rgb-d
  slam algorithm,'' in \emph{2012 IEEE/RSJ International Conference on
  Intelligent Robots and Systems}.\hskip 1em plus 0.5em minus 0.4em\relax IEEE,
  2012, pp. 1714--1719.

\bibitem{zhu2015vision}
Z.~Zhu, A.~Su, H.~Liu, Y.~Shang, and Q.~Yu, ``Vision navigation for aircrafts
  based on 3d reconstruction from real-time image sequences,'' \emph{Science
  China Technological Sciences}, vol.~58, no.~7, pp. 1196--1208, 2015.

\bibitem{chai2017obstacle}
X.~Chai, F.~Gao, C.~Qi, Y.~Pan, Y.~Xu, and Y.~Zhao, ``Obstacle avoidance for a
  hexapod robot in unknown environment,'' \emph{Science China Technological
  Sciences}, vol.~60, no.~6, pp. 818--831, 2017.

\bibitem{park2017rdfnet}
S.-J. Park, K.-S. Hong, and S.~Lee, ``Rdfnet: Rgb-d multi-level residual
  feature fusion for indoor semantic segmentation,'' in \emph{Proceedings of
  the IEEE International Conference on Computer Vision}, 2017, pp. 4980--4989.

\bibitem{ullman1979interpretation}
S.~Ullman, ``The interpretation of structure from motion,'' \emph{Proceedings
  of the Royal Society of London. Series B. Biological Sciences}, vol. 203, no.
  1153, pp. 405--426, 1979.

\bibitem{mancini2013using}
F.~Mancini, M.~Dubbini, M.~Gattelli, F.~Stecchi, S.~Fabbri, and G.~Gabbianelli,
  ``Using unmanned aerial vehicles (uav) for high-resolution reconstruction of
  topography: The structure from motion approach on coastal environments,''
  \emph{Remote Sensing}, vol.~5, no.~12, pp. 6880--6898, 2013.

\bibitem{mur2015orb}
R.~Mur-Artal, J.~M.~M. Montiel, and J.~D. Tardos, ``Orb-slam: a versatile and
  accurate monocular slam system,'' \emph{IEEE transactions on robotics},
  vol.~31, no.~5, pp. 1147--1163, 2015.

\bibitem{szeliski1997shape}
R.~Szeliski and S.~B. Kang, ``Shape ambiguities in structure from motion,''
  \emph{IEEE Transactions on Pattern Analysis and Machine Intelligence},
  vol.~19, no.~5, pp. 506--512, 1997.

\bibitem{zou2010method}
L.~Zou and Y.~Li, ``A method of stereo vision matching based on opencv,'' in
  \emph{2010 International Conference on Audio, Language and Image
  Processing}.\hskip 1em plus 0.5em minus 0.4em\relax IEEE, 2010, pp. 185--190.

\bibitem{cao2015summary}
Z.-L. Cao, Z.-H. Yan, and H.~Wang, ``Summary of binocular stereo vision
  matching technology,'' \emph{Journal of Chongqing University of Technology
  (Natural Science)}, vol.~29, no.~2, pp. 70--75, 2015.

\bibitem{benosman1996multidirectional}
R.~Benosman, T.~Mani{\`e}re, and J.~Devars, ``Multidirectional stereovision
  sensor, calibration and scenes reconstruction,'' in \emph{Proceedings of 13th
  International Conference on Pattern Recognition}, vol.~1.\hskip 1em plus
  0.5em minus 0.4em\relax IEEE, 1996, pp. 161--165.

\bibitem{ramirez2020improve}
L.~R. Ram{\'\i}rez-Hern{\'a}ndez, J.~C. Rodr{\'\i}guez-Qui{\~n}onez, M.~J.
  Castro-Toscano, D.~Hern{\'a}ndez-Balbuena, W.~Flores-Fuentes,
  R.~Rasc{\'o}n-Carmona, L.~Lindner, and O.~Sergiyenko, ``Improve
  three-dimensional point localization accuracy in stereo vision systems using
  a novel camera calibration method,'' \emph{International Journal of Advanced
  Robotic Systems}, vol.~17, no.~1, p. 1729881419896717, 2020.

\bibitem{godard2017unsupervised}
C.~Godard, O.~Mac~Aodha, and G.~J. Brostow, ``Unsupervised monocular depth
  estimation with left-right consistency,'' in \emph{Proceedings of the IEEE
  Conference on Computer Vision and Pattern Recognition}, 2017, pp. 270--279.

\bibitem{zhan2018unsupervised}
H.~Zhan, R.~Garg, C.~Saroj~Weerasekera, K.~Li, H.~Agarwal, and I.~Reid,
  ``Unsupervised learning of monocular depth estimation and visual odometry
  with deep feature reconstruction,'' in \emph{Proceedings of the IEEE
  Conference on Computer Vision and Pattern Recognition}, 2018, pp. 340--349.

\bibitem{yin2018geonet}
Z.~Yin and J.~Shi, ``Geonet: Unsupervised learning of dense depth, optical flow
  and camera pose,'' in \emph{Proceedings of the IEEE Conference on Computer
  Vision and Pattern Recognition}, 2018, pp. 1983--1992.

\bibitem{wang2018learning}
C.~Wang, J.~Miguel~Buenaposada, R.~Zhu, and S.~Lucey, ``Learning depth from
  monocular videos using direct methods,'' in \emph{Proceedings of the IEEE
  Conference on Computer Vision and Pattern Recognition}, 2018, pp. 2022--2030.

\bibitem{fei2019geo}
X.~Fei, A.~Wong, and S.~Soatto, ``Geo-supervised visual depth prediction,''
  \emph{IEEE Robotics and Automation Letters}, vol.~4, no.~2, pp. 1661--1668,
  2019.

\bibitem{tateno2017cnn}
``Cnn-slam, keisuke and tombari, federico and laina, iro and navab, nassir,''
  in \emph{Proceedings of the IEEE Conference on Computer Vision and Pattern
  Recognition}, 2017, pp. 6243--6252.

\bibitem{yoneda2014lidar}
K.~Yoneda, H.~Tehrani, T.~Ogawa, N.~Hukuyama, and S.~Mita, ``Lidar scan feature
  for localization with highly precise 3-d map,'' in \emph{2014 IEEE
  Intelligent Vehicles Symposium Proceedings}.\hskip 1em plus 0.5em minus
  0.4em\relax IEEE, 2014, pp. 1345--1350.

\bibitem{zhang2019fast}
F.~Zhang, X.~Zhu, and M.~Ye, ``Fast human pose estimation,'' in
  \emph{Proceedings of the IEEE Conference on Computer Vision and Pattern
  Recognition}, 2019, pp. 3517--3526.

\bibitem{pang2019libra}
J.~Pang, K.~Chen, J.~Shi, H.~Feng, W.~Ouyang, and D.~Lin, ``Libra r-cnn:
  Towards balanced learning for object detection,'' in \emph{Proceedings of the
  IEEE Conference on Computer Vision and Pattern Recognition}, 2019, pp.
  821--830.

\bibitem{lyu2019esnet}
H.~Lyu, H.~Fu, X.~Hu, and L.~Liu, ``Esnet: Edge-based segmentation network for
  real-time semantic segmentation in traffic scenes,'' in \emph{2019 IEEE
  International Conference on Image Processing (ICIP)}.\hskip 1em plus 0.5em
  minus 0.4em\relax IEEE, 2019, pp. 1855--1859.

\bibitem{zhao2019object}
Z.-Q. Zhao, P.~Zheng, S.-t. Xu, and X.~Wu, ``Object detection with deep
  learning: A review,'' \emph{IEEE transactions on neural networks and learning
  systems}, vol.~30, no.~11, pp. 3212--3232, 2019.

\bibitem{ghosh2019understanding}
S.~Ghosh, N.~Das, I.~Das, and U.~Maulik, ``Understanding deep learning
  techniques for image segmentation,'' \emph{ACM Computing Surveys (CSUR)},
  vol.~52, no.~4, pp. 1--35, 2019.

\bibitem{rawat2017deep}
W.~Rawat and Z.~Wang, ``Deep convolutional neural networks for image
  classification: A comprehensive review,'' \emph{Neural computation}, vol.~29,
  no.~9, pp. 2352--2449, 2017.

\bibitem{tang2020perception}
Y.~Tang, C.~Zhao, J.~Wang, C.~Zhang, Q.~Sun, and F.~Qian, ``Perception and
  decision-making of autonomous systems in the era of learning: An overview,''
  \emph{arXiv preprint arXiv:2001.02319}, 2020.

\bibitem{facil2019cam}
J.~M. Facil, B.~Ummenhofer, H.~Zhou, L.~Montesano, T.~Brox, and J.~Civera,
  ``Cam-convs: camera-aware multi-scale convolutions for single-view depth,''
  in \emph{Proceedings of the IEEE conference on computer vision and pattern
  recognition}, 2019, pp. 11\,826--11\,835.

\bibitem{garg2016unsupervised}
R.~Garg, V.~K. BG, G.~Carneiro, and I.~Reid, ``Unsupervised cnn for single view
  depth estimation: Geometry to the rescue,'' in \emph{European Conference on
  Computer Vision}.\hskip 1em plus 0.5em minus 0.4em\relax Springer, 2016, pp.
  740--756.

\bibitem{wang2019recurrent}
R.~Wang, S.~M. Pizer, and J.-M. Frahm, ``Recurrent neural network for (un-)
  supervised learning of monocular video visual odometry and depth,'' in
  \emph{Proceedings of the IEEE Conference on Computer Vision and Pattern
  Recognition}, 2019, pp. 5555--5564.

\bibitem{chakravarty2019gen}
P.~Chakravarty, P.~Narayanan, and T.~Roussel, ``Gen-slam: Generative modeling
  for monocular simultaneous localization and mapping,'' in \emph{2019
  International Conference on Robotics and Automation (ICRA)}.\hskip 1em plus
  0.5em minus 0.4em\relax IEEE, 2019, pp. 147--153.

\bibitem{aleotti2018generative}
F.~Aleotti, F.~Tosi, M.~Poggi, and S.~Mattoccia, ``Generative adversarial
  networks for unsupervised monocular depth prediction,'' in \emph{European
  Conference on Computer Vision}.\hskip 1em plus 0.5em minus 0.4em\relax
  Springer, 2018, pp. 337--354.

\bibitem{geiger2012we}
A.~Geiger, P.~Lenz, and R.~Urtasun, ``Are we ready for autonomous driving? the
  kitti vision benchmark suite,'' in \emph{2012 IEEE Conference on Computer
  Vision and Pattern Recognition}.\hskip 1em plus 0.5em minus 0.4em\relax IEEE,
  2012, pp. 3354--3361.

\bibitem{mayer2016large}
N.~Mayer, E.~Ilg, P.~Hausser, P.~Fischer, D.~Cremers, A.~Dosovitskiy, and
  T.~Brox, ``A large dataset to train convolutional networks for disparity,
  optical flow, and scene flow estimation,'' in \emph{Proceedings of the IEEE
  Conference on Computer Vision and Pattern Recognition}, 2016, pp. 4040--4048.

\bibitem{zhao2019deep}
C.~Zhao, Y.~Tang, and Q.~Sun, ``Deep direct visual odometry,'' \emph{arXiv
  preprint arXiv:1912.05101}, 2019.

\bibitem{eigen2014depth}
D.~Eigen, C.~Puhrsch, and R.~Fergus, ``Depth map prediction from a single image
  using a multi-scale deep network,'' in \emph{Advances in neural information
  processing systems}, 2014, pp. 2366--2374.

\bibitem{chen2017multi}
X.~Chen, H.~Ma, J.~Wan, B.~Li, and T.~Xia, ``Multi-view 3d object detection
  network for autonomous driving,'' in \emph{Proceedings of the IEEE Conference
  on Computer Vision and Pattern Recognition}, 2017, pp. 1907--1915.

\bibitem{wang2018understanding}
P.~Wang, P.~Chen, Y.~Yuan, D.~Liu, Z.~Huang, X.~Hou, and G.~Cottrell,
  ``Understanding convolution for semantic segmentation,'' in \emph{2018 IEEE
  winter conference on applications of computer vision (WACV)}.\hskip 1em plus
  0.5em minus 0.4em\relax IEEE, 2018, pp. 1451--1460.

\bibitem{chang2019argoverse}
M.-F. Chang, J.~Lambert, P.~Sangkloy, J.~Singh, S.~Bak, A.~Hartnett, D.~Wang,
  P.~Carr, S.~Lucey, D.~Ramanan \emph{et~al.}, ``Argoverse: 3d tracking and
  forecasting with rich maps,'' in \emph{Proceedings of the IEEE Conference on
  Computer Vision and Pattern Recognition}, 2019, pp. 8748--8757.

\bibitem{xue2019beyond}
F.~Xue, X.~Wang, S.~Li, Q.~Wang, J.~Wang, and H.~Zha, ``Beyond tracking:
  Selecting memory and refining poses for deep visual odometry,'' in
  \emph{Proceedings of the IEEE Conference on Computer Vision and Pattern
  Recognition}, 2019, pp. 8575--8583.

\bibitem{clark2017vinet}
R.~Clark, S.~Wang, H.~Wen, A.~Markham, and N.~Trigoni, ``Vinet: Visual-inertial
  odometry as a sequence-to-sequence learning problem,'' in \emph{Thirty-First
  AAAI Conference on Artificial Intelligence}, 2017.

\bibitem{silberman2012indoor}
N.~Silberman, D.~Hoiem, P.~Kohli, and R.~Fergus, ``Indoor segmentation and
  support inference from rgbd images,'' in \emph{European conference on
  computer vision}.\hskip 1em plus 0.5em minus 0.4em\relax Springer, 2012, pp.
  746--760.

\bibitem{cordts2016cityscapes}
M.~Cordts, M.~Omran, S.~Ramos, T.~Rehfeld, M.~Enzweiler, R.~Benenson,
  U.~Franke, S.~Roth, and B.~Schiele, ``The cityscapes dataset for semantic
  urban scene understanding,'' in \emph{Proceedings of the IEEE conference on
  computer vision and pattern recognition}, 2016, pp. 3213--3223.

\bibitem{zhou2017unsupervised}
T.~Zhou, M.~Brown, N.~Snavely, and D.~G. Lowe, ``Unsupervised learning of depth
  and ego-motion from video,'' in \emph{Proceedings of the IEEE Conference on
  Computer Vision and Pattern Recognition}, 2017, pp. 1851--1858.

\bibitem{bian2019unsupervised}
J.~Bian, Z.~Li, N.~Wang, H.~Zhan, C.~Shen, M.-M. Cheng, and I.~Reid,
  ``Unsupervised scale-consistent depth and ego-motion learning from monocular
  video,'' in \emph{Advances in Neural Information Processing Systems}, 2019,
  pp. 35--45.

\bibitem{saxena2008make3d}
A.~Saxena, M.~Sun, and A.~Y. Ng, ``Make3d: Learning 3d scene structure from a
  single still image,'' \emph{IEEE transactions on pattern analysis and machine
  intelligence}, vol.~31, no.~5, pp. 824--840, 2008.

\bibitem{hoiem2005automatic}
D.~Hoiem, A.~A. Efros, and M.~Hebert, ``Automatic photo pop-up,'' in \emph{ACM
  SIGGRAPH 2005 Papers}, 2005, pp. 577--584.

\bibitem{dijk2019neural}
T.~v. Dijk and G.~d. Croon, ``How do neural networks see depth in single
  images?'' in \emph{Proceedings of the IEEE International Conference on
  Computer Vision}, 2019, pp. 2183--2191.

\bibitem{kuznietsov2017semi}
Y.~Kuznietsov, J.~Stuckler, and B.~Leibe, ``Semi-supervised deep learning for
  monocular depth map prediction,'' in \emph{Proceedings of the IEEE conference
  on computer vision and pattern recognition}, 2017, pp. 6647--6655.

\bibitem{kendall2017end}
A.~Kendall, H.~Martirosyan, S.~Dasgupta, P.~Henry, R.~Kennedy, A.~Bachrach, and
  A.~Bry, ``End-to-end learning of geometry and context for deep stereo
  regression,'' in \emph{Proceedings of the IEEE International Conference on
  Computer Vision}, 2017, pp. 66--75.

\bibitem{mahjourian2018unsupervised}
R.~Mahjourian, M.~Wicke, and A.~Angelova, ``Unsupervised learning of depth and
  ego-motion from monocular video using 3d geometric constraints,'' in
  \emph{Proceedings of the IEEE Conference on Computer Vision and Pattern
  Recognition}, 2018, pp. 5667--5675.

\bibitem{eigen2015predicting}
D.~Eigen and R.~Fergus, ``Predicting depth, surface normals and semantic labels
  with a common multi-scale convolutional architecture,'' in \emph{Proceedings
  of the IEEE international conference on computer vision}, 2015, pp.
  2650--2658.

\bibitem{shelhamer2015scene}
E.~Shelhamer, J.~T. Barron, and T.~Darrell, ``Scene intrinsics and depth from a
  single image,'' in \emph{Proceedings of the IEEE International Conference on
  Computer Vision Workshops}, 2015, pp. 37--44.

\bibitem{he2016deep}
K.~He, X.~Zhang, S.~Ren, and J.~Sun, ``Deep residual learning for image
  recognition,'' in \emph{Proceedings of the IEEE conference on computer vision
  and pattern recognition}, 2016, pp. 770--778.

\bibitem{laina2016deeper}
I.~Laina, C.~Rupprecht, V.~Belagiannis, F.~Tombari, and N.~Navab, ``Deeper
  depth prediction with fully convolutional residual networks,'' in \emph{2016
  Fourth international conference on 3D vision (3DV)}.\hskip 1em plus 0.5em
  minus 0.4em\relax IEEE, 2016, pp. 239--248.

\bibitem{zwald2012berhu}
L.~Zwald and S.~Lambert-Lacroix, ``The berhu penalty and the grouped effect,''
  \emph{arXiv preprint arXiv:1207.6868}, 2012.

\bibitem{zhang2018joint}
Z.~Zhang, Z.~Cui, C.~Xu, Z.~Jie, X.~Li, and J.~Yang, ``Joint task-recursive
  learning for semantic segmentation and depth estimation,'' in
  \emph{Proceedings of the European Conference on Computer Vision (ECCV)},
  2018, pp. 235--251.

\bibitem{mancini2016fast}
M.~Mancini, G.~Costante, P.~Valigi, and T.~A. Ciarfuglia, ``Fast robust
  monocular depth estimation for obstacle detection with fully convolutional
  networks,'' in \emph{2016 IEEE/RSJ International Conference on Intelligent
  Robots and Systems (IROS)}.\hskip 1em plus 0.5em minus 0.4em\relax IEEE,
  2016, pp. 4296--4303.

\bibitem{chen2016single}
W.~Chen, Z.~Fu, D.~Yang, and J.~Deng, ``Single-image depth perception in the
  wild,'' in \emph{Advances in neural information processing systems}, 2016,
  pp. 730--738.

\bibitem{szegedy2015going}
C.~Szegedy, W.~Liu, Y.~Jia, P.~Sermanet, S.~Reed, D.~Anguelov, D.~Erhan,
  V.~Vanhoucke, and A.~Rabinovich, ``Going deeper with convolutions,'' in
  \emph{Proceedings of the IEEE conference on computer vision and pattern
  recognition}, 2015, pp. 1--9.

\bibitem{fu2018deep}
H.~Fu, M.~Gong, C.~Wang, K.~Batmanghelich, and D.~Tao, ``Deep ordinal
  regression network for monocular depth estimation,'' in \emph{Proceedings of
  the IEEE Conference on Computer Vision and Pattern Recognition}, 2018, pp.
  2002--2011.

\bibitem{wofk2019fastdepth}
D.~Wofk, F.~Ma, T.-J. Yang, S.~Karaman, and V.~Sze, ``Fastdepth: Fast monocular
  depth estimation on embedded systems,'' in \emph{2019 International
  Conference on Robotics and Automation (ICRA)}.\hskip 1em plus 0.5em minus
  0.4em\relax IEEE, 2019, pp. 6101--6108.

\bibitem{goodfellow2014generative}
I.~Goodfellow, J.~Pouget-Abadie, M.~Mirza, B.~Xu, D.~Warde-Farley, S.~Ozair,
  A.~Courville, and Y.~Bengio, ``Generative adversarial nets,'' in
  \emph{Advances in neural information processing systems}, 2014, pp.
  2672--2680.

\bibitem{jung2017depth}
H.~Jung, Y.~Kim, D.~Min, C.~Oh, and K.~Sohn, ``Depth prediction from a single
  image with conditional adversarial networks,'' in \emph{2017 IEEE
  International Conference on Image Processing (ICIP)}.\hskip 1em plus 0.5em
  minus 0.4em\relax IEEE, 2017, pp. 1717--1721.

\bibitem{cs2018monocular}
A.~CS~Kumar, S.~M. Bhandarkar, and M.~Prasad, ``Monocular depth prediction
  using generative adversarial networks,'' in \emph{Proceedings of the IEEE
  Conference on Computer Vision and Pattern Recognition Workshops}, 2018, pp.
  300--308.

\bibitem{feng2019sganvo}
T.~Feng and D.~Gu, ``Sganvo: Unsupervised deep visual odometry and depth
  estimation with stacked generative adversarial networks,'' \emph{IEEE
  Robotics and Automation Letters}, vol.~4, no.~4, pp. 4431--4437, 2019.

\bibitem{li2015depth}
B.~Li, C.~Shen, Y.~Dai, A.~Van Den~Hengel, and M.~He, ``Depth and surface
  normal estimation from monocular images using regression on deep features and
  hierarchical crfs,'' in \emph{Proceedings of the IEEE conference on computer
  vision and pattern recognition}, 2015, pp. 1119--1127.

\bibitem{huang2011hierarchical}
Q.~Huang, M.~Han, B.~Wu, and S.~Ioffe, ``A hierarchical conditional random
  field model for labeling and segmenting images of street scenes,'' in
  \emph{CVPR 2011}.\hskip 1em plus 0.5em minus 0.4em\relax IEEE, 2011, pp.
  1953--1960.

\bibitem{ladicky2009associative}
L.~Ladick{\`y}, C.~Russell, P.~Kohli, and P.~H. Torr, ``Associative
  hierarchical crfs for object class image segmentation,'' in \emph{2009 IEEE
  12th International Conference on Computer Vision}.\hskip 1em plus 0.5em minus
  0.4em\relax IEEE, 2009, pp. 739--746.

\bibitem{liu2015learning}
F.~Liu, C.~Shen, G.~Lin, and I.~Reid, ``Learning depth from single monocular
  images using deep convolutional neural fields,'' \emph{IEEE transactions on
  pattern analysis and machine intelligence}, vol.~38, no.~10, pp. 2024--2039,
  2015.

\bibitem{wang2015towards}
P.~Wang, X.~Shen, Z.~Lin, S.~Cohen, B.~Price, and A.~L. Yuille, ``Towards
  unified depth and semantic prediction from a single image,'' in
  \emph{Proceedings of the IEEE Conference on Computer Vision and Pattern
  Recognition}, 2015, pp. 2800--2809.

\bibitem{mousavian2016joint}
A.~Mousavian, H.~Pirsiavash, and J.~Ko{\v{s}}eck{\'a}, ``Joint semantic
  segmentation and depth estimation with deep convolutional networks,'' in
  \emph{2016 Fourth International Conference on 3D Vision (3DV)}.\hskip 1em
  plus 0.5em minus 0.4em\relax IEEE, 2016, pp. 611--619.

\bibitem{xu2018structured}
D.~Xu, W.~Wang, H.~Tang, H.~Liu, N.~Sebe, and E.~Ricci, ``Structured attention
  guided convolutional neural fields for monocular depth estimation,'' in
  \emph{Proceedings of the IEEE Conference on Computer Vision and Pattern
  Recognition}, 2018, pp. 3917--3925.

\bibitem{roy2016monocular}
A.~Roy and S.~Todorovic, ``Monocular depth estimation using neural regression
  forest,'' in \emph{Proceedings of the IEEE conference on computer vision and
  pattern recognition}, 2016, pp. 5506--5514.

\bibitem{zhang2017stackgan}
H.~Zhang, T.~Xu, H.~Li, S.~Zhang, X.~Wang, X.~Huang, and D.~N. Metaxas,
  ``Stackgan: Text to photo-realistic image synthesis with stacked generative
  adversarial networks,'' in \emph{Proceedings of the IEEE international
  conference on computer vision}, 2017, pp. 5907--5915.

\bibitem{hong2019generative}
Y.~Hong, U.~Hwang, J.~Yoo, and S.~Yoon, ``How generative adversarial networks
  and their variants work: An overview,'' \emph{ACM Computing Surveys (CSUR)},
  vol.~52, no.~1, pp. 1--43, 2019.

\bibitem{huang2017stacked}
X.~Huang, Y.~Li, O.~Poursaeed, J.~Hopcroft, and S.~Belongie, ``Stacked
  generative adversarial networks,'' in \emph{Proceedings of the IEEE
  conference on computer vision and pattern recognition}, 2017, pp. 5077--5086.

\bibitem{mirza2014conditional}
M.~Mirza and S.~Osindero, ``Conditional generative adversarial nets,''
  \emph{arXiv preprint arXiv:1411.1784}, 2014.

\bibitem{zhu2017unpaired}
J.-Y. Zhu, T.~Park, P.~Isola, and A.~A. Efros, ``Unpaired image-to-image
  translation using cycle-consistent adversarial networks,'' in
  \emph{Proceedings of the IEEE international conference on computer vision},
  2017, pp. 2223--2232.

\bibitem{gwn2018generative}
K.~Gwn~Lore, K.~Reddy, M.~Giering, and E.~A. Bernal, ``Generative adversarial
  networks for depth map estimation from rgb video,'' in \emph{Proceedings of
  the IEEE Conference on Computer Vision and Pattern Recognition Workshops},
  2018, pp. 1177--1185.

\bibitem{szeliski1999prediction}
R.~Szeliski, ``Prediction error as a quality metric for motion and stereo,'' in
  \emph{Proceedings of the Seventh IEEE International Conference on Computer
  Vision}, vol.~2.\hskip 1em plus 0.5em minus 0.4em\relax IEEE, 1999, pp.
  781--788.

\bibitem{godard2019digging}
C.~Godard, O.~Mac~Aodha, M.~Firman, and G.~J. Brostow, ``Digging into
  self-supervised monocular depth estimation,'' in \emph{Proceedings of the
  IEEE International Conference on Computer Vision}, 2019, pp. 3828--3838.

\bibitem{casser2019depth}
V.~Casser, S.~Pirk, R.~Mahjourian, and A.~Angelova, ``Depth prediction without
  the sensors: Leveraging structure for unsupervised learning from monocular
  videos,'' in \emph{Proceedings of the AAAI Conference on Artificial
  Intelligence}, vol.~33, 2019, pp. 8001--8008.

\bibitem{bozorgtabar2019syndemo}
B.~Bozorgtabar, M.~S. Rad, D.~Mahapatra, and J.-P. Thiran, ``Syndemo:
  Synergistic deep feature alignment for joint learning of depth and
  ego-motion,'' in \emph{Proceedings of the IEEE International Conference on
  Computer Vision}, 2019, pp. 4210--4219.

\bibitem{heise2013pm}
P.~Heise, S.~Klose, B.~Jensen, and A.~Knoll, ``Pm-huber: Patchmatch with huber
  regularization for stereo matching,'' in \emph{Proceedings of the IEEE
  International Conference on Computer Vision}, 2013, pp. 2360--2367.

\bibitem{vijayanarasimhan2017sfm}
S.~Vijayanarasimhan, S.~Ricco, C.~Schmid, R.~Sukthankar, and K.~Fragkiadaki,
  ``Sfm-net: Learning of structure and motion from video,'' \emph{arXiv
  preprint arXiv:1704.07804}, 2017.

\bibitem{yang2017unsupervised}
Z.~Yang, P.~Wang, W.~Xu, L.~Zhao, and R.~Nevatia, ``Unsupervised learning of
  geometry with edge-aware depth-normal consistency,'' \emph{arXiv preprint
  arXiv:1711.03665}, 2017.

\bibitem{wang2019unsupervised}
G.~Wang, H.~Wang, Y.~Liu, and W.~Chen, ``Unsupervised learning of monocular
  depth and ego-motion using multiple masks,'' in \emph{2019 International
  Conference on Robotics and Automation (ICRA)}.\hskip 1em plus 0.5em minus
  0.4em\relax IEEE, 2019, pp. 4724--4730.

\bibitem{sun2019cycle}
Q.~Sun, Y.~Tang, and C.~Zhao, ``Cycle-sfm: Joint self-supervised learning of
  depth and camera motion from monocular image sequences,'' \emph{Chaos: An
  Interdisciplinary Journal of Nonlinear Science}, vol.~29, no.~12, p. 123102,
  2019.

\bibitem{zou2018df}
Y.~Zou, Z.~Luo, and J.-B. Huang, ``Df-net: Unsupervised joint learning of depth
  and flow using cross-task consistency,'' in \emph{Proceedings of the European
  Conference on Computer Vision (ECCV)}, 2018, pp. 36--53.

\bibitem{ranjan2019competitive}
A.~Ranjan, V.~Jampani, L.~Balles, K.~Kim, D.~Sun, J.~Wulff, and M.~J. Black,
  ``Competitive collaboration: Joint unsupervised learning of depth, camera
  motion, optical flow and motion segmentation,'' in \emph{Proceedings of the
  IEEE Conference on Computer Vision and Pattern Recognition}, 2019, pp.
  12\,240--12\,249.

\bibitem{chen2019self}
Y.~Chen, C.~Schmid, and C.~Sminchisescu, ``Self-supervised learning with
  geometric constraints in monocular video: Connecting flow, depth, and
  camera,'' in \emph{Proceedings of the IEEE International Conference on
  Computer Vision}, 2019, pp. 7063--7072.

\bibitem{gordon2019depth}
A.~Gordon, H.~Li, R.~Jonschkowski, and A.~Angelova, ``Depth from videos in the
  wild: Unsupervised monocular depth learning from unknown cameras,'' in
  \emph{Proceedings of the IEEE International Conference on Computer Vision},
  2019, pp. 8977--8986.

\bibitem{li2019sequential}
S.~Li, F.~Xue, X.~Wang, Z.~Yan, and H.~Zha, ``Sequential adversarial learning
  for self-supervised deep visual odometry,'' in \emph{Proceedings of the IEEE
  International Conference on Computer Vision}, 2019, pp. 2851--2860.

\bibitem{almalioglu2019ganvo}
Y.~Almalioglu, M.~R.~U. Saputra, P.~P. de~Gusmao, A.~Markham, and N.~Trigoni,
  ``Ganvo: Unsupervised deep monocular visual odometry and depth estimation
  with generative adversarial networks,'' in \emph{2019 International
  Conference on Robotics and Automation (ICRA)}.\hskip 1em plus 0.5em minus
  0.4em\relax IEEE, 2019, pp. 5474--5480.

\bibitem{wang2004image}
Z.~Wang, A.~C. Bovik, H.~R. Sheikh, and E.~P. Simoncelli, ``Image quality
  assessment: from error visibility to structural similarity,'' \emph{IEEE
  transactions on image processing}, vol.~13, no.~4, pp. 600--612, 2004.

\bibitem{poggi2018learning}
M.~Poggi, F.~Tosi, and S.~Mattoccia, ``Learning monocular depth estimation with
  unsupervised trinocular assumptions,'' in \emph{2018 International Conference
  on 3D Vision (3DV)}.\hskip 1em plus 0.5em minus 0.4em\relax IEEE, 2018, pp.
  324--333.

\bibitem{ramirez2018geometry}
P.~Z. Ramirez, M.~Poggi, F.~Tosi, S.~Mattoccia, and L.~Di~Stefano, ``Geometry
  meets semantics for semi-supervised monocular depth estimation,'' in
  \emph{Asian Conference on Computer Vision}.\hskip 1em plus 0.5em minus
  0.4em\relax Springer, 2018, pp. 298--313.

\bibitem{chen2019towards}
P.-Y. Chen, A.~H. Liu, Y.-C. Liu, and Y.-C.~F. Wang, ``Towards scene
  understanding: Unsupervised monocular depth estimation with semantic-aware
  representation,'' in \emph{Proceedings of the IEEE Conference on Computer
  Vision and Pattern Recognition}, 2019, pp. 2624--2632.

\bibitem{luo2018single}
Y.~Luo, J.~Ren, M.~Lin, J.~Pang, W.~Sun, H.~Li, and L.~Lin, ``Single view
  stereo matching,'' in \emph{Proceedings of the IEEE Conference on Computer
  Vision and Pattern Recognition}, 2018, pp. 155--163.

\bibitem{xie2016deep3d}
J.~Xie, R.~Girshick, and A.~Farhadi, ``Deep3d: Fully automatic 2d-to-3d video
  conversion with deep convolutional neural networks,'' in \emph{European
  Conference on Computer Vision}.\hskip 1em plus 0.5em minus 0.4em\relax
  Springer, 2016, pp. 842--857.

\bibitem{tosi2019learning}
F.~Tosi, F.~Aleotti, M.~Poggi, and S.~Mattoccia, ``Learning monocular depth
  estimation infusing traditional stereo knowledge,'' in \emph{Proceedings of
  the IEEE Conference on Computer Vision and Pattern Recognition}, 2019, pp.
  9799--9809.

\bibitem{pilzer2018unsupervised}
A.~Pilzer, D.~Xu, M.~Puscas, E.~Ricci, and N.~Sebe, ``Unsupervised adversarial
  depth estimation using cycled generative networks,'' in \emph{2018
  International Conference on 3D Vision (3DV)}.\hskip 1em plus 0.5em minus
  0.4em\relax IEEE, 2018, pp. 587--595.

\bibitem{pilzer2019refine}
A.~Pilzer, S.~Lathuiliere, N.~Sebe, and E.~Ricci, ``Refine and distill:
  Exploiting cycle-inconsistency and knowledge distillation for unsupervised
  monocular depth estimation,'' in \emph{Proceedings of the IEEE Conference on
  Computer Vision and Pattern Recognition}, 2019, pp. 9768--9777.

\bibitem{zhao2019geometry}
S.~Zhao, H.~Fu, M.~Gong, and D.~Tao, ``Geometry-aware symmetric domain
  adaptation for monocular depth estimation,'' in \emph{Proceedings of the IEEE
  Conference on Computer Vision and Pattern Recognition}, 2019, pp. 9788--9798.

\bibitem{wu2019spatial}
Z.~Wu, X.~Wu, X.~Zhang, S.~Wang, and L.~Ju, ``Spatial correspondence with
  generative adversarial network: Learning depth from monocular videos,'' in
  \emph{Proceedings of the IEEE International Conference on Computer Vision},
  2019, pp. 7494--7504.

\bibitem{he2018wearable}
L.~He, C.~Chen, T.~Zhang, H.~Zhu, and S.~Wan, ``Wearable depth camera:
  Monocular depth estimation via sparse optimization under weak supervision,''
  \emph{IEEE Access}, vol.~6, pp. 41\,337--41\,345, 2018.

\bibitem{loo2019cnn}
S.~Y. Loo, A.~J. Amiri, S.~Mashohor, S.~H. Tang, and H.~Zhang, ``Cnn-svo:
  Improving the mapping in semi-direct visual odometry using single-image depth
  prediction,'' in \emph{2019 International Conference on Robotics and
  Automation (ICRA)}.\hskip 1em plus 0.5em minus 0.4em\relax IEEE, 2019, pp.
  5218--5223.

\bibitem{forster2014svo}
C.~Forster, M.~Pizzoli, and D.~Scaramuzza, ``Svo: Fast semi-direct monocular
  visual odometry,'' in \emph{2014 IEEE international conference on robotics
  and automation (ICRA)}.\hskip 1em plus 0.5em minus 0.4em\relax IEEE, 2014,
  pp. 15--22.

\bibitem{engel2017direct}
J.~Engel, V.~Koltun, and D.~Cremers, ``Direct sparse odometry,'' \emph{IEEE
  transactions on pattern analysis and machine intelligence}, vol.~40, no.~3,
  pp. 611--625, 2017.

\bibitem{yin2017scale}
X.~Yin, X.~Wang, X.~Du, and Q.~Chen, ``Scale recovery for monocular visual
  odometry using depth estimated with deep convolutional neural fields,'' in
  \emph{Proceedings of the IEEE International Conference on Computer Vision},
  2017, pp. 5870--5878.

\bibitem{yang2018deep}
N.~Yang, R.~Wang, J.~Stuckler, and D.~Cremers, ``Deep virtual stereo odometry:
  Leveraging deep depth prediction for monocular direct sparse odometry,'' in
  \emph{Proceedings of the European Conference on Computer Vision (ECCV)},
  2018, pp. 817--833.

\bibitem{li2018undeepvo}
R.~Li, S.~Wang, Z.~Long, and D.~Gu, ``Undeepvo: Monocular visual odometry
  through unsupervised deep learning,'' in \emph{2018 IEEE international
  conference on robotics and automation (ICRA)}.\hskip 1em plus 0.5em minus
  0.4em\relax IEEE, 2018, pp. 7286--7291.

\bibitem{wang2019unos}
Y.~Wang, P.~Wang, Z.~Yang, C.~Luo, Y.~Yang, and W.~Xu, ``Unos: Unified
  unsupervised optical-flow and stereo-depth estimation by watching videos,''
  in \emph{Proceedings of the IEEE Conference on Computer Vision and Pattern
  Recognition}, 2019, pp. 8071--8081.

\bibitem{watson2019self}
J.~Watson, M.~Firman, G.~J. Brostow, and D.~Turmukhambetov, ``Self-supervised
  monocular depth hints,'' in \emph{Proceedings of the IEEE International
  Conference on Computer Vision}, 2019, pp. 2162--2171.

\bibitem{cheng2018learning}
X.~Cheng, P.~Wang, and R.~Yang, ``Learning depth with convolutional spatial
  propagation network,'' \emph{arXiv preprint arXiv:1810.02695}, 2018.

\bibitem{li2018deeper}
Q.~Li, Z.~Han, and X.-M. Wu, ``Deeper insights into graph convolutional
  networks for semi-supervised learning,'' in \emph{Thirty-Second AAAI
  Conference on Artificial Intelligence}, 2018.

\bibitem{vaswani2017attention}
A.~Vaswani, N.~Shazeer, N.~Parmar, J.~Uszkoreit, L.~Jones, A.~N. Gomez,
  {\L}.~Kaiser, and I.~Polosukhin, ``Attention is all you need,'' in
  \emph{Advances in neural information processing systems}, 2017, pp.
  5998--6008.

\bibitem{ahn2019variational}
S.~Ahn, S.~X. Hu, A.~Damianou, N.~D. Lawrence, and Z.~Dai, ``Variational
  information distillation for knowledge transfer,'' in \emph{Proceedings of
  the IEEE Conference on Computer Vision and Pattern Recognition}, 2019, pp.
  9163--9171.

\bibitem{wu2016stability}
X.~Wu, Y.~Tang, and W.~Zhang, ``Stability analysis of stochastic delayed
  systems with an application to multi-agent systems,'' \emph{IEEE Transactions
  on Automatic Control}, vol.~61, no.~12, pp. 4143--4149, 2016.

\bibitem{tang2020input}
Y.~Tang, X.~Wu, P.~Shi, and F.~Qian, ``Input-to-state stability for nonlinear
  systems with stochastic impulses,'' \emph{Automatica}, vol. 113, p. 108766,
  2020.

\end{thebibliography}
}

\end{document}